\documentclass{article}

\usepackage[final]{neurips_data_2022}

\usepackage[utf8]{inputenc} 
\usepackage[T1]{fontenc}    
\usepackage{hyperref}       
\usepackage{url}            
\usepackage{booktabs}       
\usepackage{amsfonts}       
\usepackage{nicefrac}       
\usepackage{microtype}      
\usepackage{xcolor}         
\usepackage{enumitem}

\usepackage{microtype}
\usepackage{multirow}
\usepackage{graphicx}
\usepackage{subfigure}
\usepackage{booktabs} 
\usepackage[normalem]{ulem}

\usepackage{hyperref}

\usepackage{booktabs}       
\usepackage{amsfonts}       
\usepackage{nicefrac}       
\usepackage{microtype}      
\usepackage{xcolor}         

\usepackage{rotating}

\usepackage{amsmath}
\usepackage{amsfonts}
\usepackage{amssymb}
\usepackage{amsthm}
\usepackage{graphicx}
\usepackage{color}
\usepackage{url}
\usepackage{stackrel}
\usepackage{subfigure}
\usepackage{xspace}
\usepackage{tabularx}
\usepackage{balance}
\usepackage{wrapfig}
\usepackage{multirow}
\usepackage{booktabs}
\usepackage{tabu}
\usepackage[linesnumbered,ruled,noend]{algorithm2e}
\usepackage{subfigure}
\usepackage{graphicx}
\usepackage{float}

\theoremstyle{remark}

\newcommand{\data}{\mathcal{D}\xspace}

\newcommand{\features}{\mathcal{X}\xspace}
\newcommand{\labels}{\mathcal{Y}\xspace}

\newcommand{\ie}{{\em i.e.}\xspace}
\newcommand{\eg}{{\em e.g.}\xspace}

\newcommand{\squishlist}{
	\begin{list}{$\bullet$}{
			\setlength{\itemsep}{0pt}
			\setlength{\parsep}{3pt}
			\setlength{\topsep}{3pt}
			\setlength{\partopsep}{0pt}
			\setlength{\leftmargin}{1.0em}
			\setlength{\labelwidth}{1em}
			\setlength{\labelsep}{0.5em}
		}
	}
	
	\newcommand{\squishenum}{
		
		\begin{list}{\usecounter{scount}}{
				\setlength{\itemsep}{0pt}
				\setlength{\parsep}{3pt}
				\setlength{\topsep}{3pt}
				\setlength{\partopsep}{0pt}
				\setlength{\leftmargin}{1.2em}
				\setlength{\labelwidth}{1em}
				\setlength{\labelsep}{0.5em}
			}
		}
		
		\newcommand{\squishend}{
		\end{list}
	}

\usepackage{amsmath}
\usepackage{amssymb}
\usepackage{mathtools}
\usepackage{amsthm}

\usepackage[capitalize,noabbrev]{cleveref}

\theoremstyle{definition}

\usepackage[textsize=tiny]{todonotes}

\usepackage{amsmath,amsfonts,bm}

\makeatletter
\def\widebreve{\mathpalette\wide@breve}
\def\wide@breve#1#2{\sbox\z@{$#1#2$}%
	\mathop{\vbox{\m@th\ialign{##\crcr
				\kern0.08em\brevefill#1{0.6\wd\z@}\crcr\noalign{\nointerlineskip}%
				$\hss#1#2\hss$\crcr}}}\limits}
\def\brevefill#1#2{$\m@th\sbox\tw@{$#1($}%
	\hss\resizebox{#2}{\wd\tw@}{\rotatebox[origin=c]{90}{\upshape(}}\hss$}
\makeatletter

\def\1{\bm{1}}

\DeclareMathAlphabet{\mathsfit}{\encodingdefault}{\sfdefault}{m}{sl}
\SetMathAlphabet{\mathsfit}{bold}{\encodingdefault}{\sfdefault}{bx}{n}

\title{pFL-Bench: A Comprehensive Benchmark for Personalized Federated Learning}

\author{%
  Daoyuan Chen, Dawei Gao, Weirui Kuang, Yaliang Li\thanks{corresponding author},~ Bolin Ding \\
  Alibaba Group \\
  \texttt{\{daoyuanchen.cdy, gaodawei.gdw, weirui.kwr\}@alibaba-inc.com} \\
  \texttt{\{yaliang.li, bolin.ding\}@alibaba-inc.com} \\
}

\begin{document}

\maketitle

\begin{abstract}
  Personalized Federated Learning (pFL), which utilizes and deploys distinct local models, has gained increasing attention in recent years due to its success in handling the statistical heterogeneity of FL clients.
  However, standardized evaluation and systematical analysis of diverse pFL methods remain a challenge. Firstly, the highly varied datasets, FL simulation settings and pFL implementations prevent easy and fair comparisons of pFL methods. Secondly, the current pFL literature diverges in the adopted evaluation and ablation protocols. Finally, the effectiveness and robustness of pFL methods are under-explored in various practical scenarios, such as the generalization to new clients and the participation of resource-limited clients. To tackle these challenges, we propose the first comprehensive pFL benchmark, pFL-Bench, for facilitating rapid, reproducible, standardized and thorough pFL evaluation. The proposed benchmark contains more than 10 dataset variants in various application domains with a unified data partition and realistic heterogeneous settings; a modularized and easy-to-extend pFL codebase with more than 20 competitive pFL method implementations; and systematic evaluations under containerized environments in terms of generalization, fairness, system overhead, and convergence.
  We highlight the benefits and potential of state-of-the-art pFL methods and hope the pFL-Bench enables further pFL research and broad applications that would otherwise be difficult owing to the absence of a dedicated benchmark.
  The code is released at https://github.com/alibaba/FederatedScope/tree/master/benchmark/pFL-Bench. \footnote[1]{We will continuously maintain the benchmark and update the codebase and arXiv version.}
\end{abstract}

\section{Introduction}
\label{sec:intro}

Federated learning (FL) is an emerging machine learning (ML) paradigm, which collaboratively trains models via coordinating certain distributed clients (\eg, smart IoT devices) with a logically centralized aggregator \cite{mcmahan2017communication,hong2021federated}. 
Due to the benefit that it does not transmit local data and circumvents the high cost and privacy risks of collecting raw sensitive data from clients, FL has gained widespread interest and has been applied in numerous ML tasks such as image classification \cite{zhao2018federated,oh2022fedbabu}, object detection \cite{liu2020fedvision,Benyuan21partialFed}, keyboard suggestion \cite{hard2018federated,yang2018applied}, text  classification \cite{zhu2020empirical}, relation extraction \cite{sui2020feded}, speech recognition \cite{paulik2021federated,guliani2021training}, graph classification \cite{xie2021federated}, recommendation \cite{muhammad2020fedfast,wu2021fedgnn}, and healthcare \cite{yang2021flop,dayan2021federated}.

Pioneering FL researchers made great efforts to find a global model that performs well for most FL clients \cite{kairouz2021advances,meng2021cross,yu2021fed2,mothukuri2021survey}.
However, the intrinsic statistical and system heterogeneity of clients limits the performance and applicability of such classical FL methods \cite{Kulkarni2020SurveyOP,tan2021towards}. 
Taking the concept shift case as an example, in which the
conditional distribution $P(Y|X)$ vary across some clients who have the same marginal distributions $P(X)$ \cite{kairouz2021advances}, a shared global model cannot fit these clients well at the same time. 
Recently, the personalized FL (pFL) methods that utilize client-distinct models to overcome these challenges have been gaining increasing popularity, such as those based on multi-task learning \cite{Liam2021Shared,ghoshEfficientFrameworkClustered2020a,li2021ditto}, meta-learning \cite{Dinh2020PersonalizedFL,fallahPersonalizedFederatedLearning2020}, and transfer learning
\cite{Yang2019FederatedML,karimireddy2020scaffold,zhuDataFreeKnowledgeDistillation2021}.

Even though fruitful pFL methods have been explored, a standard benchmark is still lacking.
As a result, the evaluation of pFL methods is currently with non-standardized datasets and implementations,  highly diverse evaluation protocols, and unclear effectiveness and robustness of pFL methods under various practical scenarios. To be specific:
\begin{itemize}[leftmargin=*]
	\item \textbf{Non-standardized datasets and implementations for pFL.} 
	Currently, researchers often use custom FL datasets and implementations to evaluate the effectiveness of proposed methods due to the absence of standardized pFL benchmarks. 
	For example, although many pFL works use the same public FEMNIST \cite{caldas2018leaf} and CIFAR-10/100 \cite{Krizhevsky09learningmultiple} datasets, the partition manners can be divergent: the number of clients is 205 in \cite{li2021ditto} while 539 in \cite{fedem} for FEMNIST; and \cite{shamsian2021pFedHN} adopts the Dirichlet distribution based partition while \cite{zhang2020personalized} uses the pathological partition for CIFAR-10/100. 
	Prior pFL studies set up different computation environments and simulation settings, increasing the difficulty of fast evaluation and the risk of unfair comparisons.
	\item \textbf{Diverse evaluation protocols.} 
	The current pFL methods often focus on different views and adopt diverse evaluation protocols, which may lead to isolated development of pFL and prevent pFL research from reaching its full potential. For example, besides the global accuracy improvement, a few works studied the local accuracy evaluation characterized by fairness, and system efficiency in terms of communication and computational costs \cite{li2021ditto,diaoHeteroFLComputationCommunication2020}.
	Without a careful design and control of the evaluation, it is difficult to compare the pros and cons of different pFL methods and understand how much costs we pay for the personalization.
	\item \textbf{Under-explored practicability of pFL in various scenarios.} 
	Most existing pFL methods examine their effectiveness in several mild Non-IID FL cases \cite{tan2021towards}. However, it is unclear whether existing pFL methods can consistently work well in more practical scenarios, such as the participation of partial clients in which the clients have spotty connectivity \cite{fedscale-icml}; the participation of resource-limited clients in which the personalization is required to be highly efficient \cite{oort-osdi21}; and the generalization to new clients in which learned models will be applied to new clients that do not participate in the FL process \cite{yuan2022what}.
\end{itemize}
To quantify the progress in the pFL community and facilitate rapid, reproducible, and generalizable pFL research, we propose the first comprehensive pFL benchmark characterized as follows:
\begin{itemize}[leftmargin=*]
	\item We provide 4 benchmark families with 12 dataset variants for diverse application domains involving image, text, graph and recommendation data, each with unified data partition and some realistic Non-IID settings such as clients sampling and the participation of new clients. Some public popular DataZoos such as LEAF \cite{caldas2018leaf}, Torchvision \cite{torchvision} and Huggingface datasets \cite{lhoest-etal-2021-datasets} are also compatible to the proposed pFL-Bench to enable flexible and easily-extended experiments.
	\item We implement an extensible open-sourced pFL codebase that contains more than 20 pFL methods, providing fruitful state-of-the-art (SOTA) methods re-implementations, unified interfaces and pluggable personalization sub-modules such as model parameter decoupling, model mixture, meta-learning and personalized regularization. 
	\item We conduct systematic evaluation under unified experimental settings and containerized environments to show the efficacy of pFL-Bench and provide standardized scripts to ensure the reproducibility and maintainability of pFL-Bench.
	We also highlight the advantages of pFL methods and opportunities for further pFL study in terms of generalization, fairness, system overhead and convergence.
\end{itemize}

\section{Related Works}
\paragraph{Personalized Federated Learning.}
Despite the promising performance using a shared global model for all clients as demonstrated in \cite{mcmahan2017communication,FedOPT2020Asad,FedNova2020Wang,karimireddy2019scaffold,Sahu2018OnTC,huang2021fl}, it is challenging to find a converged best-for-all global model under statistical and system heterogeneity among clients \cite{sattler2020clustered,li2020federated,chai2019towards}.
As a natural way to handle the heterogeneity,
personalization is gaining popularity in recent years.
Fruitful pFL literatures have explored the accuracy and convergence improvement based on clustering \cite{briggs2020federated,sattler2020clustered,chai2020tifl},  multi-task learning \cite{smith2017federated,corinzia2019variational,huang2021personalized,marfoq2021fedEM}, model mixture \cite{zhang2020personalized,li2021ditto,zhang2021personalized,hanzely2020lower}, model parameter decoupling \cite{diaoHeteroFLComputationCommunication2020,Benyuan21partialFed}, Bayesian treatment \cite{achituve2021personalized,corinzia2019variational}, knowledge distillation \cite{linEnsembleDistillationRobust2021,zhuDataFreeKnowledgeDistillation2021,ozkara2021quped}, meta-learning \cite{khodak2019adaptive,jiangImprovingFederatedLearning2019,khodak2019adaptive,fallahPersonalizedFederatedLearning2020,singhal2021federated,acar2021debiasing}, and transfer learning \cite{yangFedStegFederatedTransfer2020,annavaramGroupKnowledgeTransfer,zhang2021parameterized}. 
We refer readers to related FL and pFL survey paper for more details \cite{kairouz2021advances,Kulkarni2020SurveyOP,tan2021towards}.
In pFL-Bench, we provide modularized re-implementation for numerous SOTA pFL methods with several fundamental and pluggable sub-routines for easy and fast pFL research and deployment. 
We plan to add more pFL methods in the future and also welcome contributions to the pFL-Bench.

\paragraph{Federated Learning Benchmark.}
We are aware that there are great efforts on benchmarking FL from various aspects, such as heterogeneous datasets (LEAF \cite{caldas2018leaf}, TFF \cite{tff}), heterogeneous system resources (Flower \cite{flower}, FedML \cite{fedml}, FedScale \cite{fedscale-icml}), and specific domains (FedNLP \cite{fednlp2021}, FS-G \cite{wang2022federatedscopegnn}). 
However, they mostly benchmarked general FL algorithms, lacking recently proposed pFL methods that perform well on heterogeneous FL scenarios.
Besides, no benchmark so far supports the evaluation of generalization to new clients; and few existing benchmark simultaneously supports comprehensive evaluation for trade-offs among accuracy, fairness and system costs. We hope to close these gaps with this proposed pFL-Bench, and facilitate further pFL research and broad applications. 

\section{Background and Problem Formulation}
\subsection{A Generalized FL Setting}
We first introduce some important concepts in FL, taking the FedAvg method \cite{mcmahan2017communication} as an illustrative example. 
A typical FL procedure using FedAvg is as follows: 
Each client $i \in \mathcal{C}$ has its own private dataset $\data_i$ over $\features \times \labels$, and the goal of FL is to train a single global model $\theta_g$ with collaborative learning from this set of clients $\mathcal{C}$ without directly sharing their local data.
At each FL round, the server broadcasts $\theta_g$ to selected clients $\mathcal{C}_s \subseteq \mathcal{C}$, who then perform local learning based on the private local data and upload the local update information (\eg, the gradients of trained models) to the server.
After collecting and averagely aggregating the update information from clients, the server applies the updates into $\theta_g$ for next-round federation and the process repeats.

Then we present a generalized FL formulation, which establishes the proposed comprehensive benchmark in terms of diverse evaluation and personalization perspectives. Specifically, besides the FL-participated clients $\mathcal{C}$, we consider a set of new clients that do not participate to the FL training process and denote it as $\mathcal{\tilde{C}}$. Most FL approaches implicitly solve the following problem:
\begin{equation}
\label{eq:overall-obj}
\begin{split}
\min_{\{h_{\theta_g}\}  \cup \{h_{\theta_i}\}_{i \in \mathcal{C}} \cup \{h_{\theta_j}\}_{j \in \tilde{\mathcal{C}}}
} 
\alpha  G\big( \{
\mathbb{E}_{(x, y) \sim \data_i} & [f(\theta_g; x, y)] \}_{i \in \mathcal{C}} \big) + 
\beta L\big( \{
 \mathbb{E}_{(x, y) \sim \data_i} [f(\theta_i; x, y)] \}_{i\in\mathcal{C}} \big) \\ 
+ \gamma R\big( \{
\mathbb{E}_{(x, y) \sim \data_j} & [f(\theta_j; x, y) |\theta_g] \}_{j\in\mathcal{\tilde{C}}} \big) + \zeta Q(\theta_g, \theta_k) _{k \in (\mathcal{C} \cup \tilde{\mathcal{C}})}, \qquad 
\end{split}
\end{equation}
where $f(\theta; x, y)$ indicates the loss at data point $(x, y)$ with model $\theta$. 
The term $G(\cdot)$ indicates the global objective based on the shared global model $\theta_g=Agg([\theta_i]_{i\in\mathcal{C}})$ with an aggregation function $Agg(\cdot)$ for model parameters, and
$L(\cdot)$ indicates the local objective based on the local distinct models $[\theta_i]_{i\in\mathcal{C}}$. 
We note that $G(\cdot)$ and $L(\cdot)$ can be in various forms such as uniform averaging or weighted averaging according to local training data size of clients, which corresponds to the commonly used intra-client generalization case \cite{yuan2022what}.

For the latter two terms, $R(\cdot)$ usually has similar forms to $G(\cdot)$ and measures the generalization to new clients $\mathcal{\tilde{C}}$ that do not contribute to the FL training process. $Q(\cdot)$ indicates the modeling of the relationship between the global and local models, such as the $L^2$ norm to regularize the model parameters in Ditto \cite{li2021ditto}. 
Besides, different pFL methods may flexibly introduce various constraints on this optimization objective, which we have omitted in Eq.\eqref{eq:overall-obj} for brevity.
The coefficients $\alpha$, $\beta$, $\gamma$ and $\zeta$ trade off these terms.
For non-personalized FL algorithms, $\beta=0$.
Although the generalization term $R(\cdot)$ has been primarily explored in a few recent studies \cite{yuan2022what,shamsian2021pFedHN}, most existing FL works overlooked it with $\gamma=0$.
Later, we will discuss more instantiations for $L(\cdot)$ and $Q(\cdot)$ in the personalization setting.

\subsection{Personalization Setup}
With the above generalized formulation, we can see that existing pFL works achieve personalization via multi-granularity objects, including the global model $\theta_{g}$ and local models $[\theta_{i}]$. For example, many two-step pFL approaches first find a strong $\theta_{g}$ in the FL training stages, then get local models $[\theta_{i}]$ by fine-tuning $\theta_{g}$ on local data and use $[\theta_{i}]$ in inference \cite{tan2021towards,mansour2020three}.
A more flexible manner is to directly learn distinct local models $[\theta_{i}]$ in the FL process, while this introduces additional storage and computation costs for clients \cite{deng2020adaptive}.
Recently, to gain a better accuracy-efficiency trade-off, several pFL works propose to only personalize a sub-module $\pi_i$ of $\theta_i$, and transmit and aggregate the remaining part as $\theta_g=Agg([\theta_i \setminus \pi_i]_{i \in \mathcal{C}})$ \cite{Liam2021Shared}.

We illustrate some representative personalization operations in pFL works w.r.t. the different choices of the local objective $L(\cdot)$ and $Q(\cdot)$.
Fine-tuning is a basic step widely used in abundant pFL works \cite{jiangImprovingFederatedLearning2019} to minimize
$\mathbb{E}_{(x, y) \sim \mathcal{D}_i} [f(\theta_i; x, y)]$, where $\theta_i$ is usually initialized using the parameters of $\theta_g$ before fine-tuning.
Model mixture is a general pFL approach assuming that the local data distribution is a mixture of $K$ underlying data distributions 
$\data_i=Mix(\data_{i,k}, w_{i,k})$ with mixture weight $w_{i,k}$ $\text{for}~ k\in[K]$ 
\cite{marfoq2021fedEM}, thus learning a group of intermediate models is suitable to handle the data heterogeneity as
$\theta_i=Mix(\theta_{i,k})_{k\in[K]}$.
For clustering-based pFL methods, $K$ indicates the cluster number, and the mixture weight is 0-1 indicative function for belonged clusters \cite{ghoshEfficientFrameworkClustered2020a}.
Besides, taking $\theta_g$ as the reference point, model interpolation 
$\theta_i \equiv w_g\theta_{g} + (1-w_g)\theta_{i}
$ and model regularization $\theta_i=
 argmin\big(\sum_{(x, y) \sim \data_i} (f(\theta_i; x, y)+\frac{\lambda}{2}||\theta_i-\theta_g||)\big)
$ are also widely explored in the pFL literature \cite{hanzely2020federated}, where $\lambda$ is the regularization factor.

\section{Benchmark Design and Resources}
\subsection{Datasets and Models}
\begin{figure}[t]
	\centering
		\begin{minipage}[]{\linewidth}
			\centering
			\includegraphics[width=0.99\linewidth]{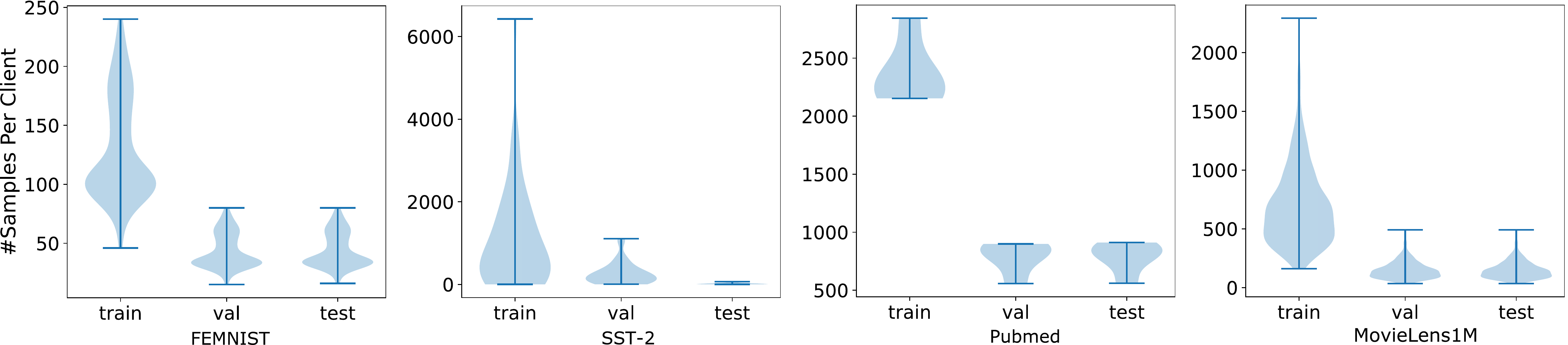}
			\caption{The violin plot of number of samples per client for partial adopted datasets. In Appendix \ref{append:data-detail}, we present the plots for other datasets, the label skew visualization and  clients’ pairwise similarity of label distribution in terms of Jensen–Shannon distance.}
	\label{fig:client-statis}
		\end{minipage}%
\end{figure}

We conduct experiments on 12 publicly available dataset variants  with heterogeneous partition in our benchmark.
These datasets are popular in the corresponding fields, and cover a wide range of domains, scales, partition manners and Non-IID degrees. We list the statistics in Table \ref{table:data-statis} and illustrate the violin plot of data size per client in Figure \ref{fig:client-statis}, which show diverse properties across the FL datasets, enabling thorough comparisons among different pFL methods.
We provide more detailed descriptions of these datasets in the Appendix \ref{append:data-detail}.
Besides, with a carefully designed modularity, our code-base is compatible with a large number of datasets from other public popular DataZoos,  including LEAF~\cite{caldas2018leaf}, Torchvision \cite{torchvision}, Huggingface datasets \cite{lhoest-etal-2021-datasets} and FederatedScope-GNN \cite{wang2022federatedscopegnn}.

\begin{table}
	\centering
	\small
	\caption{Statistics of the experimental datasets, tasks, and models in pFL-bench. We sample 5\% clients from FEMNIST \cite{caldas2018leaf}. 
		Following previous works \cite{fedml,marfoq2021fedEM,diaoHeteroFLComputationCommunication2020,fallahPersonalizedFederatedLearning2020}, we adopt Dirichlet allocation with different $\alpha$s to simulate the heterogeneous partition for CIFAR10 and textual datasets. The $\mu$ and $\sigma$ indicate the mean and std of number of samples per client.  More datasets from popular Datazoos such as LEAF~\cite{caldas2018leaf}, Torchvision \cite{torchvision}, Huggingface datasets \cite{lhoest-etal-2021-datasets} and FederatedScope-GNN \cite{wang2022federatedscopegnn} are also supported. Detailed descriptions can be found in Appendix 
		\ref{append:data-detail}.
		}
	\label{table:data-statis}
	\begin{tabular}{lcccccc}
		\toprule
		Dataset & Task & Model & Partition By & \# Clients & \multicolumn{2}{c}{\# Sample Per Client} \\ 
		\midrule
		FEMNIST & \multirow{4}{*}{Image Classification} & \multirow{4}{*}{CNN} & Writers & 200 & $\mu$=217& $\sigma$=73 \\
		CIFAR10-$\alpha5$ &  && \multirow{3}{*}{Labels} &100 & $\mu$=600& $\sigma$=46 \\
		CIFAR10-$\alpha0.5$ && && 100 & $\mu$=600& $\sigma$=137 \\
		CIFAR10-$\alpha0.1$ & & && 100 & $\mu$=600& $\sigma$=383 \\ 
		\midrule
		COLA & Linguistic Acceptability & \multirow{2}{*}{BERT} & \multirow{2}{*}{Labels} &50 & $\mu$=192& $\sigma$=159 \\
		SST-2 & Sentiment Analysis &  & &50 & $\mu$=1,364& $\sigma$=1,291 \\
		Twitter & Sentiment Analysis & LR & Users & 13,203 & $\mu$=10& $\sigma$=11 \\
		\midrule
		Cora & \multirow{3}{*}{Node Classification} & \multirow{3}{*}{GIN}&\multirow{3}{*}{ Community} & 5 & $\mu$=542& $\sigma$=30 \\
		Pubmed & & & & 5 & $\mu$=3,943& $\sigma$=34 \\
		Citeseer & & & & 5 & $\mu$=665& $\sigma$=29 \\
		\midrule
		MovieLens1M & \multirow{2}{*}{Recommendation} & \multirow{2}{*}{MF} & Users & 1000 & $\mu$=1,000& $\sigma$=482 \\
		MovieLens10M &  & & Items & 1000 & $\mu$=10,000& $\sigma$=8,155 \\
		\bottomrule
	\end{tabular}
\end{table}

We preset the widely adopted CNN model \cite{reddi2021adaptive,dinh2020personalized,liang2020think,shamsian2021personalized} with additional batch normalization layers for image datasets, and the pre-trained BERT-Tiny model from \cite{turc2019} and linear regression (LR) model for the textual datasets.
For the graph and recommendation datasets, we preset the graph isomorphism neural network, GIN \cite{xu2018how}, and Matrix Factorization (MF) \cite{koren2009matrix} respectively. 
It is worth noting that pFL-Bench provides a unified model interface decoupled with FL algorithms, enabling users to easily register and use more customized models or built-in models from existing ModelZoos including Torchvision \cite{torchvision}, Huggingface \cite{wolf-etal-2020-transformers}, and FederatedScope \cite{xie2021federated}.

\paragraph{Benchmark scenarios.}
\textsc{(Generalization)}
For a comprehensive evaluation, pFL-Bench supports examining the generalization performance for both FL-participated and FL-non-participated clients, \ie, the $R(\cdot)$ term in formulation \eqref{eq:overall-obj}. Specifically, we randomly select 20\% clients as non-participated clients for each dataset, and these clients will not transmit their training-related message during the FL processes.

\textsc{(Client Sampling)} In addition to generalization performance, we also care about how the pFL methods perform when adopting client sampling in FL processes, which is useful in cross-device scenarios where a large number of clients have spotty connectivity. 
For the image, text and recommendation datasets, we uniformly sample 20\% clients without replacement from the participating clients at each FL round.

\textsc{(Cross-silo v.s. cross-device)} Note that the adopted datasets have quite different numbers of clients after heterogeneous partition.
We choose the three graph datasets with small number of clients to simulate cross-silo FL scenarios \cite{wang2022federatedscopegnn,huang2021personalized}, while the other datasets correspond to different scales of cross-device FL scenarios.  

\subsection{Methods}
We consider abundant methods for extensive pFL comparisons.
The pFL-bench provides unified and modularized interfaces for a range of popular and SOTA methods in the following three categories:
\textbf{Non-pFL methods.} As two naive methods sitting on opposite ends of the local-global spectrum, we evaluate the \emph{Global-Train} method that trains the model from centralized data merged from all clients, and \emph{Isolated} method that trains a separated models for each client without any information transmission among clients. 
Besides, we consider the classical \emph{FedAvg} \cite{mcmahan2017communication} with weighted averaging based on local data size, the \emph{FedProx} \cite{blocal} that introduces proximal term during the local training process, and \emph{FedOpt} \cite{FedOPT2020Asad} that generalizes FedAvg by introducing an optimizer for the FL server.

\textbf{pFL methods.} We compare several SOTA methods including \emph{FedBN} \cite{li2020fedbn} that is a simple yet effective method to deal with feature shift Non-IID, via locally maintaining the clients' batch normalization parameters without transmitting and aggregation;
\emph{Ditto} \cite{li2021ditto} that improves fairness and robustness of FL by training local personalized model and global model simultaneously, in which the local model update is based on regularization to global model parameters;
\emph{pFedMe} \cite{dinh2020personalized} that is a meta-learning based method and decouples personalized model and global model with Moreau envelops;
\emph{HypCluster} \cite{mansour2020three} that splits users into clusters and learns different models for different clusters;
\emph{FedEM} \cite{marfoq2021fedEM} that assumes local data distribution is a mixture of unknown underlying distributions, and correspondingly learns a mixture of multiple intermediate models with Expectation-Maximization algorithm.

\textbf{Combined variants}. Note that pFL-Bench provides pluggable re-implementations of existing methods, enabling users can pick different personalized behaviors and different personalized objects to form a new pFL variant.
Here we combine \emph{FedBN}, \emph{FedOpt}, and \emph{Fine-tuning (FT)} with other compatible methods, and provide fine-grained ablations via more than 20 method variants for systematic pFL study and explorations of pFL potential. More details of the considered methods are in Appendix \ref{append:method-detail}.

\subsection{Evaluation Criteria}
We propose a unified and comprehensive evaluation protocol in pFL-Bench, in which evaluations from multiple perspectives are taken into consideration.
(1) For the \textbf{generalization} examination, we support monitoring on both the server and client sides, with various and extensible metric aggregation manners, and a wide range of metrics for diverse tasks such as classification, regression and ranking.
(2) We also report several \textbf{fairness-related} metrics, including standard deviation, and the top and bottom deciles of performance across different clients.
(3) Numerous \textbf{systematical metrics} are considered as well, including the process running time, the memory cost w.r.t. average and peak memory usage, the total computational cost w.r.t. FLOPs in server and clients, the communication cost w.r.t. the total number of downloaded/uploaded bytes, and the number of FL rounds to convergence. Detailed description can be found in Appendix \ref{append:method-detail}.

\subsection{Codebase}
To facilitate the innovation for pFL community, our pFL-Bench contains a well-modularized, easy-to-extend codebase for standardization of implementation, evaluation, and ablation of pFL methods.

\paragraph{pFL implementations.}
We build the pFL-bench upon an event-driven FL framework \textsc{FederatedScope} (\textsc{FS}) \cite{federatedscope}, which abstracts FL information transmitting and processing as message passing and several pluggable subroutines. 
We eliminate the cumbersome engineering for coordinating FL participants with the help of \textsc{FS}, and customize many message handlers and subroutines, such as model parameter decoupling, model mixture, local fine-tuning, meta-learning and regularization for personalization.
By combining these useful and pluggable components, we re-implement a number of SOTA pFL methods with unified and extensible interfaces.
This modularity also makes the usage of pFL methods convenient, and makes the contribution of new pFL methods easy and flexible.
We release the codes with Apache License 2.0 and will continuously include more pFL methods.

\paragraph{Reproducibility.} 
To enable easily reproducible research, we conduct experiments in containerized environments and provide standardized and documented evaluation procedures for prescribed metrics. 
For fair comparisons, we search the optimal hyper-parameters using the validation sets for all methods, with early stopping and large number of total FL rounds. 
We run experiments 3 times with the optimal configurations and report the average results.
All the experiments are conducted on a cluster of  8 Tesla V100 and 64 NVIDIA GTX 1080 Ti GPUs, taking $\sim$13,000 runs with a total of $\sim$112 days process computing time.
More details, such as hyper-parameter search spaces, can be found in Appendix \ref{append:implement-detail}.
\section{Experimental Results and Analysis}
\label{sec:exp}
To demonstrate the utility of pFL-Bench in providing fair, comprehensive, and rigorous comparisons among pFL methods, we conduct extensive experiments and present some main results in terms of generalization, fairness and efficiency under various FL datasets and scenarios.
The complete experimental results are presented in Appendix D due to the space limitation.

\begin{table}[!t]
	\centering
	
	\caption{Accuracy results for both participated clients and un-participated clients. $\overline{Acc}$ indicates the aggregated accuracy weighted by the number of local data samples of participated clients, $\widetilde{Acc}$ indicates the aggregated accuracy of un-participated clients, and $\Delta$ indicates the participation generalization gap. \textbf{Bold} and \underline{underlined} indicate the best and second-best results among all compared methods, while \textcolor{red}{red} and \textcolor{blue}{blue} indicate the best and second-best results for original methods without combination ``-''.
	}
	\small
	\resizebox{\columnwidth}{!}{
		\begin{tabular}{p{1.42in}|ccc|ccc|ccc}
			\toprule
			& \multicolumn{3}{c|}{FEMNIST, $s=0.2$} & \multicolumn{3}{c|}{SST-2} & \multicolumn{3}{c}{PUBMED}  \\
			& $\overline{Acc}$ & $\widetilde{Acc}$ & $\Delta$ & $\overline{Acc}$ & $\widetilde{Acc}$ & $\Delta$& $\overline{Acc}$ & $\widetilde{Acc}$ & $\Delta$ \\
			\midrule
			Global-Train & 74.51 & - & - & \color{red}{\textbf{80.57}} & - & - & 87.01 & - & - \\
			Isolated & 68.74 & - & - & 60.82 & - & - & 85.56 & - & - \\
			FedAvg & 83.97 & 81.97 & -2.00 & 74.88 & \color{red}{80.24} & \color{red}{5.36} & \color{blue}{87.27} & \color{blue}{72.63} & -14.64 \\
			FedAvg-FT & 86.44 & 84.94 & \underline{-1.50} & 74.14 & \textbf{83.28} & 9.13 & 87.21 & \underline{79.78} & \underline{-7.43} \\
			FedProx&84.10&81.49&-2.61&74.36&79.20&4.84&87.23&75.02&-12.21\\
FedProx-FT&87.34&85.27&-2.08&79.94&80.48&0.59&\underline{88.24}&79.12&-9.12\\
			\midrule
			pFedMe & \color{blue}{87.50} & \color{blue}{82.76} & -4.73 & 71.27 & 69.34 & -1.92 & 86.91 & 71.64 & -15.27 \\
			pFedMe-FT & 88.19 & 82.46 & -5.73 & 75.61 & 66.48 & -9.13 & 85.71 & 77.07 & -8.64 \\
			\midrule
HypCluster&83.80&81.88&\color{blue}{-1.92}&46.26&61.32&15.05&87.20&\color{red}{75.37}&\color{red}{-11.83}\\
HypCluster-FT&87.79&85.67&-2.12&52.46&78.67&26.21&86.43&76.69&-9.74\\

            \midrule
			FedBN & 86.72 & 7.86 & -78.86 & 74.88 & \color{blue}{75.40} & \color{blue}{0.52} & \color{red}{\textbf{88.49}} & 52.53 & -35.95 \\
			FedBN-FT & 88.51 & 82.87 & -5.64 & 68.81 & \underline{82.43} & 13.63 & 87.45 & \textbf{80.36} & \textbf{-7.09} \\
			FedBN-FedOPT & 88.25 & 8.77 & -79.49 & 64.70 & 65.50 & 0.81 & 87.87 & 42.72 & -45.15 \\
			FedBN-FedOPT-FT & 88.14 & 80.25 & -7.88 & 68.65 & 70.56 & 1.91 & 87.54 & 77.07 & -10.47 \\
			\midrule
			Ditto & \color{red}{88.39} & 2.20 & -86.19 & 52.03 & 46.79 & -5.24 & \color{blue}{87.27} & 2.84 & -84.43 \\
			Ditto-FT & 85.72 & 56.96 & -28.76 & 56.49 & 65.50 & 9.01 & 87.47 & 35.03 & -52.44 \\
			Ditto-FedBN & \textbf{88.94} & 2.20 & -86.74 & 56.03 & 46.79 & -9.24 & 88.18 & 2.84 & -85.34 \\
			Ditto-FedBN-FT & 86.53 & 58.96 & -27.57 & 53.15 & 66.49 & 13.34 & 87.83 & 28.52 & -59.30 \\
			Ditto-FedBN-FedOpt & \underline{88.73} & 2.20 & -86.54 & 57.67 & 46.79 & -10.88 & 87.81 & 2.84 & -84.97 \\
			Ditto-FedBN-FedOpt-FT & 87.02 & 55.22 & -31.80 & 52.89 & 66.49 & 13.60 & 87.60 & 18.18 & -69.42 \\
			\midrule
			FedEM & 84.35 & \color{red}{82.81} & \color{red}{-1.54} & \color{blue}{\underline{75.78}} & 67.67 & -8.11 & 85.64 & 71.12 & \color{blue}{-14.52} \\
			FedEM-FT & 86.17 & 85.01 & \textbf{-1.16} & 64.86 & 81.63 & \textbf{16.77} & 85.88 & 78.08 & -7.80 \\
			FedEM-FedBN & 84.37 & 12.88 & -71.49 & 75.43 & 62.81 & -12.62 & 88.12 & 48.64 & -39.48 \\
			FedEM-FedBN-FT & 88.29 & 83.96 & -4.33 & 64.96 & 81.04 & \underline{16.08} & 86.38 & 72.02 & -14.35 \\
			FedEM-FedBN-FedOPT & 82.12 & 6.64 & -75.48 & 72.25 & 64.69 & -7.56 & 87.56 & 42.37 & -45.19 \\
			FedEM-FedBN-FedOPT-FT & 87.54 & \textbf{\underline{85.76}} & -1.79 & 62.26 & 73.87 & 11.61 & 87.49 & 72.39 & -15.09 \\
			\bottomrule
	\end{tabular}
	}
	\label{tab:general}
\end{table}

\subsection{Generalization}
We present the accuracy results for both participated and un-participated clients in Table \ref{tab:general} on FEMNIST, SST-2 and PUBMED datasets, where $\overline{Acc}$ and $\widetilde{Acc}$ 
indicates the aggregated accuracy weighted by the local data samples of participated and un-participated clients respectively, and $\Delta=\widetilde{Acc}-\overline{Acc}$ indicates the generalization gap.

\paragraph{Comparison of original methods.}
For the original methods without combination (``-''), we mark the best and second-best results as red and blue respectively in Table \ref{tab:general}. 
Notably, we find that \textit{no method can consistently beat others across all metrics and all datasets}. 
The pFL methods gain significantly better $\overline{Acc}$ over FedAvg in some cases (\eg, Ditto on FEMNIST), showing the effectiveness of personalization.
However, the advantages of pFL methods on un-participated clients' generalization $\widetilde{Acc}$  are relatively smaller than those on intra-client generalization $\overline{Acc}$. 
The methods Ditto and FedBN even fail to gain reasonable $\widetilde{Acc}$ on FEMNIST and PUBMED datasets, as these methods did not discuss how to apply to unseen clients, we have kept the behaviors of their algorithms in inference in order to respect the original method. 
And we examine their running dynamics and find that their local models diverge with the un-participated clients' data.
Besides, there are still performance gaps between pFL methods and the Global-Train method on the textual dataset. 
These observations demonstrate plenty of room for pFL improvement.

\paragraph{Comparison of pFL variants.}
We then extend the comparison to include FL variant methods incorporating other compatible personalized operators or methods.
In a nutshell, almost all the best and second-best results (marked as \textbf{Bold} and \underline{underlined}) are achieved by the combined variants, showing the efficacy and flexibility of our benchmark, and the potential of new pFL methods. 
Among these methods, fine-tuning (FT) effectively improves both $\overline{Acc}$ and $\widetilde{Acc}$ metrics in most cases, even for pFL methods that have been implicitly adopted local training process, which calls for deeper understanding about how much should we fit the local data and how the personalized fitting impacts the FL dynamics.
Besides, FedBN is a simple yet effective method to improve $\overline{Acc}$ while may bring negative effect on $\widetilde{Acc}$ (\eg, FedEM v.s. FedEM-FT), since the feature space of un-participated clients lack informative characterization and the frozen BN parameters of global model can arbitrarily mis-match these un-participated clients.
We also find that FedBN is much more effective on graph data with the GIN model than those on textual data with the BERT model, in which we made a simple modification that filters out the Layer Normalization parameters, showing the opportunity of designing domain- and model-specific pFL techniques.

\begin{figure}[!h]
	\centering
	\subfigure{
		\begin{minipage}[]{\linewidth}
			\centering
			\includegraphics[width=0.8\linewidth]{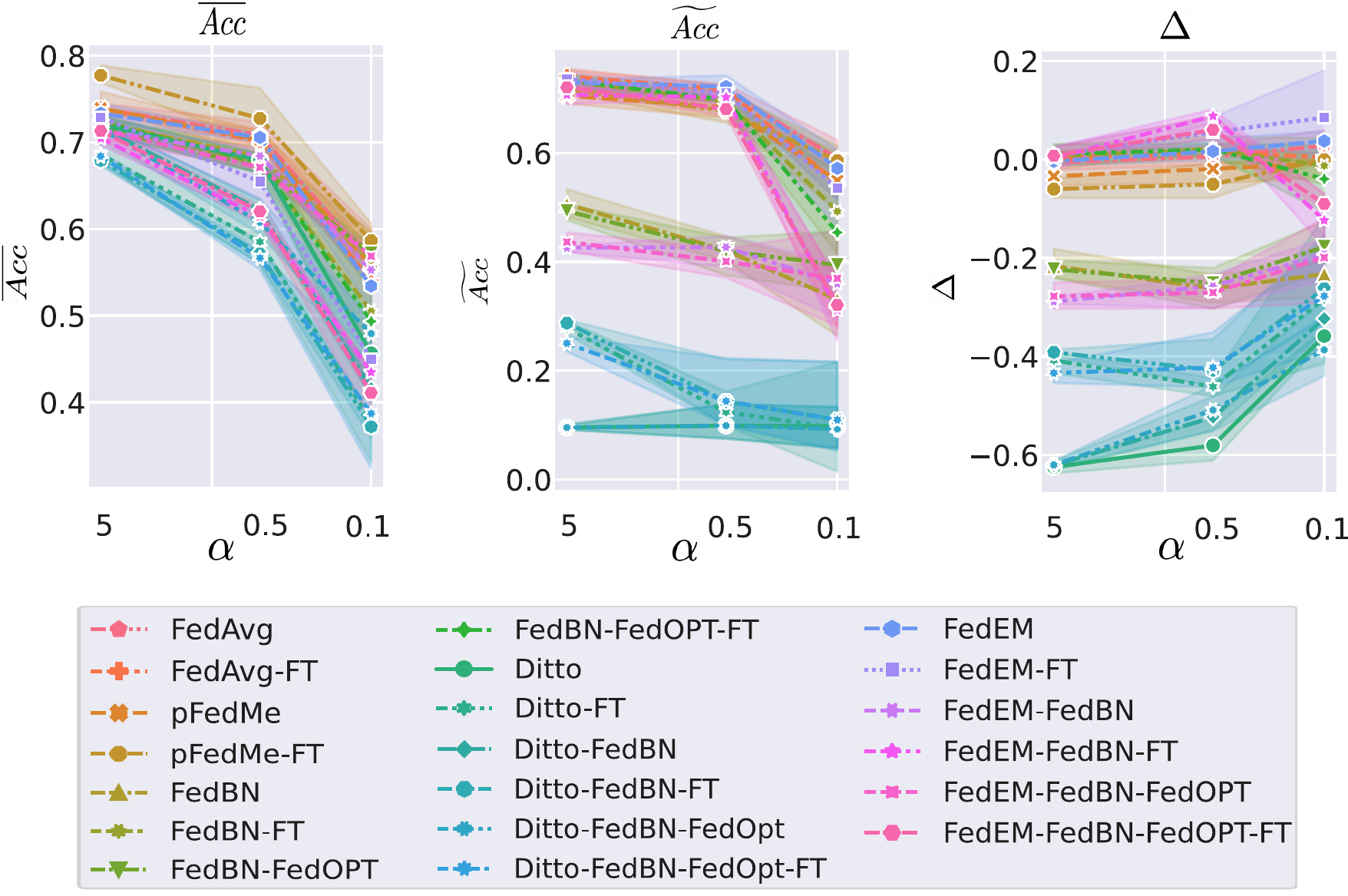}
			\caption{The accuracy results of participated, un-participated clients and  generalization gap when varying Dirichlet factor $\alpha$ for CIFAR10 dataset. 
				Results in table format are in Appendix \ref{append:general-results}.}
			\label{fig:vary_alpha}
		\end{minipage}%
	}
\end{figure}

\paragraph{Effect of Non-IID split.}
To gain further insight into the pFL methods, we vary the Non-IID degree with different Dirichlet factor $\alpha$ for CIFAR10 dataset and illustrate the results in Figure \ref{fig:vary_alpha}. 
Generally speaking, almost all methods gain performance degradation as the Non-IID degree increases (from $\alpha$=5 to $\alpha$=0.1). 
Besides, we can see that most of pFL methods show superior accuracy and robustness over FedAvg especially for the highly heterogeneous case with $\alpha$=0.1. 
These results indicate the benefit of pFL methods in Non-IID situations, as well as their substantial space for improvement.

\begin{figure}[!h]
	\centering
	\subfigure{
		\begin{minipage}[]{\linewidth}
			\centering
			\includegraphics[width=0.8\linewidth]{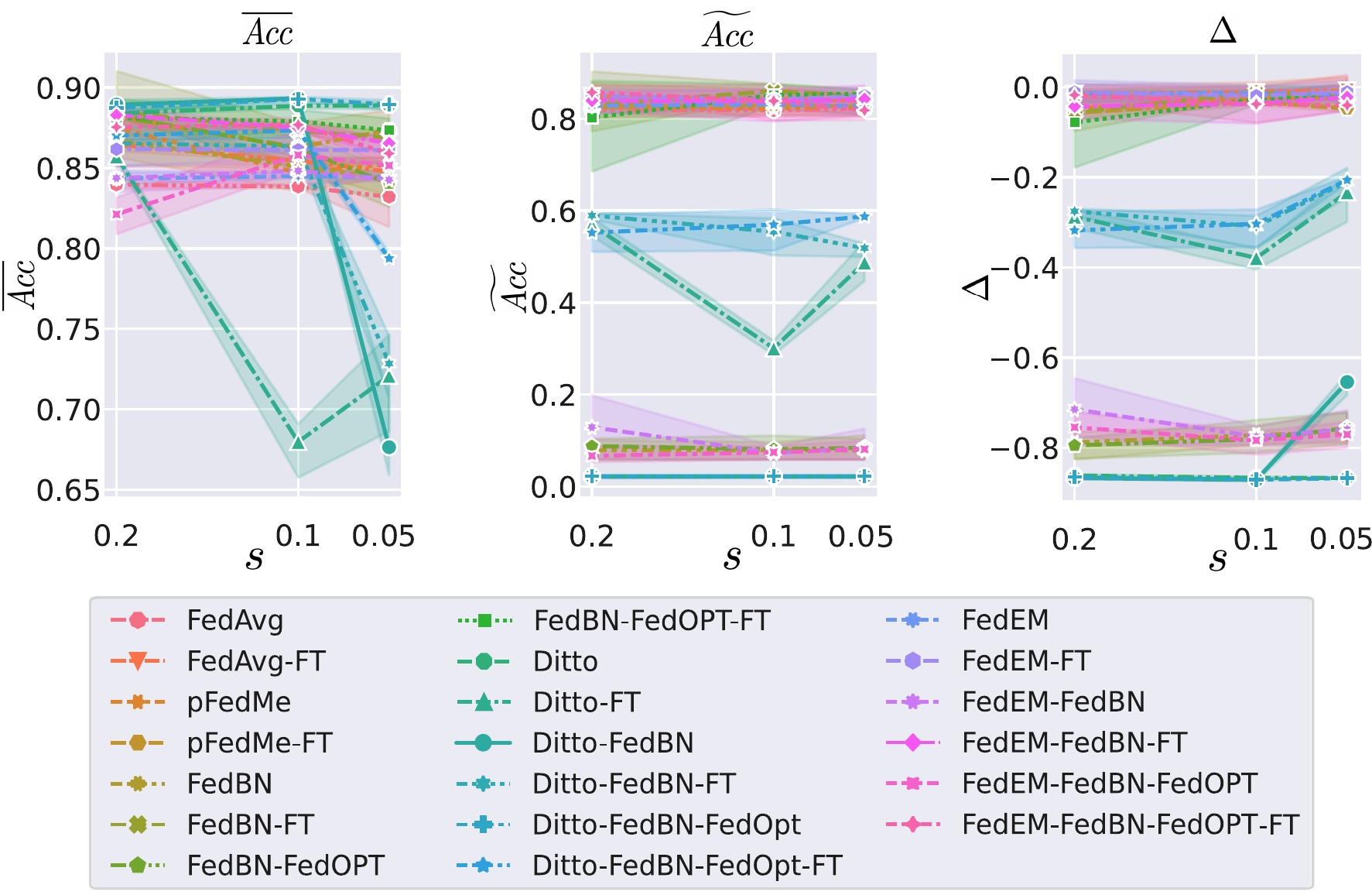}
			\caption{
				The accuracy results of participated, un-participated clients and  generalization gap when varying the client sampling rate $s$ on FEMNIST dataset. Performance of some pFL methods degrade as $s$ decreases such as Ditto and Ditto-FT, calling for robustness improvements of pFL.
				Detailed results in table format can be found in Appendix \ref{append:general-results}.
			}
			\label{fig:vary_s}
		\end{minipage}%
	}
\end{figure}

\paragraph{Effect of client sampling.}
We also vary the client sampling rate $s$ and present the generalization results on FEMNIST dataset in Figure \ref{fig:vary_s}. 
In summary, most pFL methods still achieve better results than FedAvg as $s$ decreases. 
However, the advantages are diminishing for un-participating clients and several pFL methods prune to fail with small $s$.
In addition, we observe that FT increases the performance variance in client sampling cases. 
Although several pFL works provide convergence guarantees under mild assumptions, there remain open questions about the theoretical impact of clients with spotty connections \cite{fedbuffer} in the personalization case, and the design of robust pFL algorithms.

\begin{table}[!h]
	\centering
	\small
	\caption{Fairness results in terms of $\overline{Acc}'$ indicating the equally-weighted average, $\sigma$ indicating the standard deviation of the average accuracy, and $\protect\widebreve{Acc}$ indicating the bottom accuracy. \textbf{Bold}, \underline{underlined}, \textcolor{red}{red} and \textcolor{blue}{blue} indicate the same highlights as used in Table \ref{tab:general}.}
	\resizebox{\columnwidth}{!}{
		\begin{tabular}{l|ccc|ccc|ccc}
			\toprule
			& \multicolumn{3}{c|}{FEMNIST, $s=0.2$} & \multicolumn{3}{c|}{SST-2} & \multicolumn{3}{c}{PUBMED} \\
			& $\overline{Acc}'$ & $\sigma$ & $\protect\widebreve{Acc}$ &  $\overline{Acc}'$ & $\sigma$ & $\protect\widebreve{Acc}$ & $\overline{Acc}'$ & $\sigma$ & $\protect\widebreve{Acc}$ \\
			\midrule
			Isolated & 67.08 & 10.76 & 53.16 & 59.40 & 41.29 & 0.00 & 84.67 & 6.26 & 74.63 \\
			FedAvg & 82.40 & 9.91 & 69.11 & \color{blue}{\underline{76.30}} & \color{red}{\underline{22.02}} & \color{red}{\underline{44.85}} & 86.72 & 3.93 & 79.76 \\
			FedAvg-FT & 85.17 & 8.69 & 72.34 & 75.36 & 27.67 & 31.08 & 86.71 & 3.86 & 80.57 \\
			FedProx&82.37&10.25&68.57&72.25&22.03&43.51&\textbf{\color{red}{88.06}}&\color{blue}{3.62}&80.08\\
FedProx-FT&86.06&7.89&74.60&58.72&37.58&4.17&87.87&4.02&78.80\\
\midrule
			pFedMe & \color{blue}{86.50} & 8.52 & \color{blue}{75.00} & 65.08 & 26.59 & 27.75 & 86.35 & 4.43 & 78.76 \\
			pFedMe-FT & 87.06 & 8.02 & 75.00 & 74.36 & 27.02 & 32.49 & 85.47 & \textbf{3.06} & 80.95 \\
			\midrule
			HypCluster&82.34&9.72&68.57&57.29&39.27&0.00&86.82&5.06&77.66\\
HypCluster-FT&86.58&7.84&75.71&56.47&42.59&0.00&85.97&5.69&76.45\\
\midrule
			FedBN & 85.38 & \color{blue}{8.19} & 74.26 & \color{blue}{\underline{76.30}} & \color{red}{\underline{22.02}} & \color{red}{\underline{44.85}} & \color{blue}{\underline{87.97}} & \color{red}{\underline{3.42}} & \color{red}{81.77} \\
			FedBN-FT & \underline{87.65} & \textbf{6.33} & \textbf{80.02} & 68.50 & 26.83 & 29.17 & 87.02 & 3.47 & 80.13 \\
			FedBN-FedOPT & 87.27 & 7.34 & 76.87 & 65.59 & 31.07 & 22.22 & 87.43 & 4.64 & 80.81 \\
			FedBN-FedOPT-FT & 87.13 & 7.36 & \underline{78.27} & 68.42 & 28.18 & 30.71 & 87.02 & 3.94 & 81.78 \\
			\midrule
			Ditto & \color{red}{87.18} & \color{red}{7.52} & \color{red}{78.23} & 49.94 & 40.81 & 0.00 & 86.85 & 3.98 & \color{blue}{80.44} \\
			Ditto-FT & 84.30 & 8.16 & 73.95 & 54.34 & 39.26 & 0.00 & 87.10 & 3.52 & 80.46 \\
			Ditto-FedBN & \textbf{87.82} & \underline{7.19} & 77.78 & 49.44 & 41.80 & 0.00 & 87.75 & 3.70 & 81.82 \\
			Ditto-FedBN-FT & 85.16 & 7.98 & 75.25 & 52.18 & 39.85 & 0.00 & 87.43 & 3.77 & 81.15 \\
			Ditto-FedBN-FedOpt & 87.64 & 7.08 & 78.23 & 55.61 & 40.43 & 1.39 & 87.27 & 3.90 & 79.14 \\
			Ditto-FedBN-FedOpt-FT & 85.71 & 7.91 & 75.81 & 53.16 & 34.75 & 9.72 & 87.10 & 3.79 & 80.93 \\
			\midrule
			FedEM & 82.61 & 9.57 & 69.29 & \color{red}{\textbf{76.53}} & \color{blue}{23.34} & \color{blue}{44.44} & 85.05 & 4.44 & 78.51 \\
			FedEM-FT & 84.91 & 8.39 & 73.64 & 64.29 & 32.84 & 12.96 & 85.54 & 4.48 & 79.39 \\
			FedEM-FedBN & 82.94 & 9.35 & 70.43 & 75.06 & \textbf{18.48} & \textbf{53.33} & 87.63 & 4.14 & \textbf{82.54} \\
			FedEM-FedBN-FT & 87.09 & 9.24 & 76.36 & 64.33 & 35.72 & 8.59 & 85.68 & 4.33 & 79.44 \\
			FedEM-FedBN-FedOPT & 80.48 & 11.02 & 64.84 & 72.66 & 27.18 & 34.17 & 87.11 & 4.24 & 80.32 \\
			FedEM-FedBN-FedOPT-FT & 86.23 & 8.33 & 75.44 & 58.42 & 31.21 & 17.93 & 87.16 & 3.66 & \underline{82.20}\\
			\bottomrule
	\end{tabular}}
	\label{tab:fair}
\end{table}

\subsection{Fairness Study}

We then empirically investigate what degrees of fairness can be achieved by pFL methods, and report the equally-weighted average of accuracy $\overline{Acc}'$, the standard deviation $\sigma$ across evaluated clients, and the bottom individual accuracy $\protect\widebreve{Acc}$ in Table \ref{tab:fair}. 
These metrics are considered as the fairness criteria in related pFL works \cite{li2021ditto,marfoq2021fedEM}.
We find that $\overline{Acc}'$ is usually smaller than the one weighted by local data size ($\overline{Acc}$ in Table \ref{tab:general}), indicating the client bias in existing pFL evaluation.
Across the three datasets, the $\sigma$s on SST-2 are much larger (at dozen-scales) than those on the image dataset (6.33 $\sim$ 11.02), which are larger than those on the graph dataset (3.06 $\sim$ 6.26), leaving room for further research to understand this difference and improve the fairness in various application domains.
Interestingly, compared with FedAvg, pFL methods can effectively improve bottom accuracy, while they may gain larger standard deviations. 
An exception is Ditto-based methods on SST-2, as the parameter regularization in Ditto may fail for the complex BERT model. 
Besides, the Isolated method performs bad for clients having a very small data size (a few dozens), even gains $\protect\widebreve{Acc}$=0 on SST-2. 
We find that most of the other methods achieve much better results than it, verifying the benefits of transmitting knowledge across clients.

\begin{figure}[!h]
	\centering
	\subfigure{
		\begin{minipage}[]{\linewidth}
			\centering
			\includegraphics[width=0.735\linewidth]{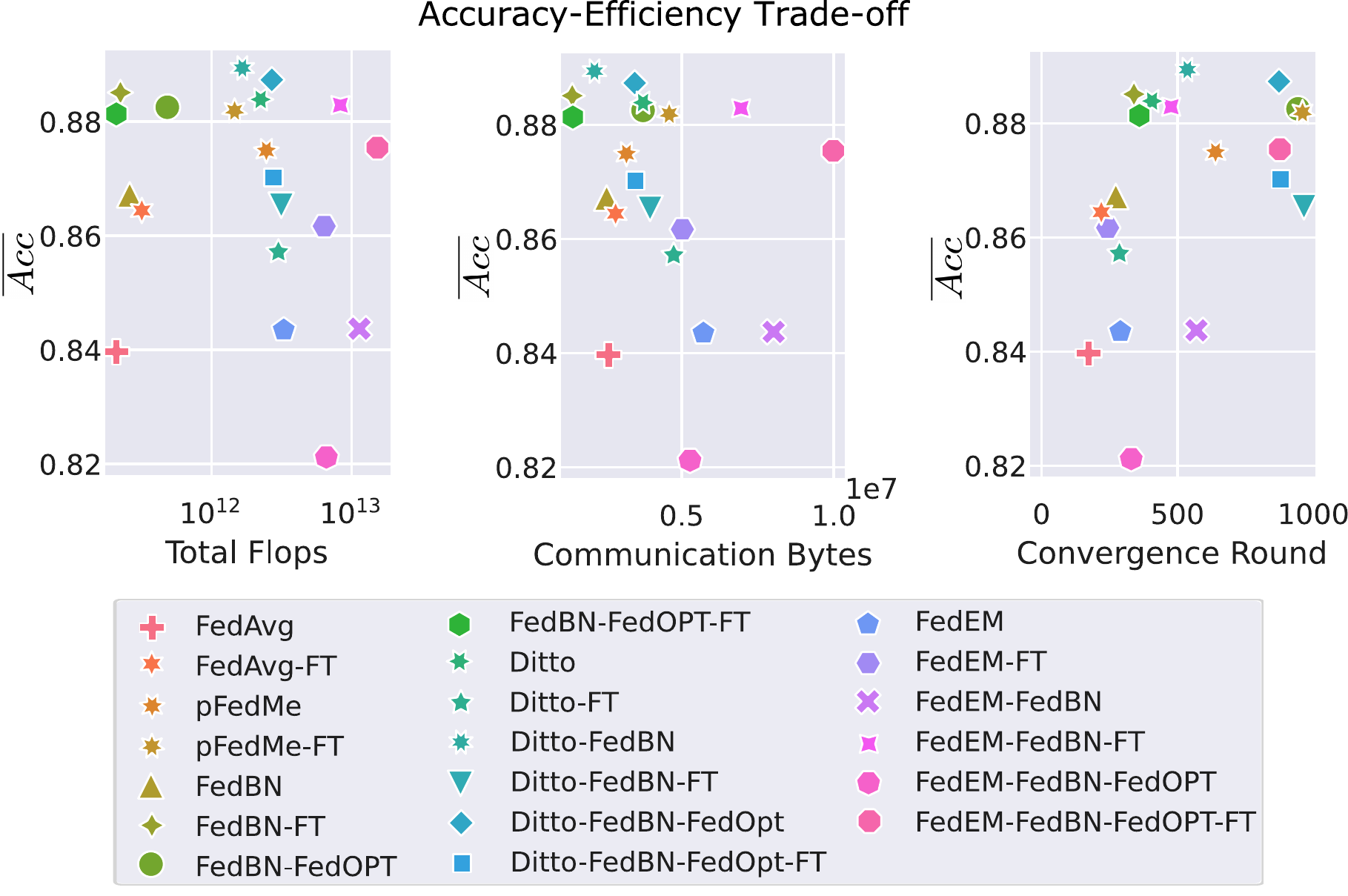}
			\caption{The trade-off between accuracy and efficiency metrics on FEMNIST dataset with client sampling rate $s=0.2$. The pFL methods usually incur much larger computation and communication costs than non-personalized methods. Full results for other datasets are shown in Appendix \ref{append:efficiency-results}. 
			}
			\label{fig:efficiency-acc-trade-off}
		\end{minipage}%
	}
\end{figure}

\subsection{Efficiency}
To quantify the systematical payloads of personalization that is introduced into the FL process, we count the total FLOPs, communication bytes and convergence rounds to demonstrate the trade-off between these metrics and accuracy in Figure \ref{fig:efficiency-acc-trade-off}. 
Not surprisingly, pFL methods usually incur much larger computation and communication costs than non-personalized methods, requiring more careful and efficient design for further pFL research in resource-limited scenarios.
Another interesting observation is that when combined with FT or FedOpt that aggregates the clients' model updates as pseudo-gradients for the global model, the convergence speeds are improved for some methods such as FedEM and Ditto, showing the potential of co-optimizing from the server and local client sides.

\subsection{More Experiments}
Due to the space limitation, we present more experimental results in Appendix \ref{append:more-exp-results}, including results in terms of generalization (Sec.\ref{append:general-results}), fairness (Sec.\ref{append:fair-results}) and efficiency (Sec.\ref{append:efficiency-results}) for all the datasets in Table \ref{table:data-statis}.
To demonstrate the potential and ease of extensibility of the pFL-bench, we also conduct experiments in the scenario of heterogeneous device resource based on FedScale \cite{fedscale-icml} in Sec.\ref{append:hetero-device}, where we adopt the over-selection mechanism for server and a temporal event simulator \cite{overselection}.
The simulator executes the behaviors of clients according to virtual timestamps of their message delivery to the server, and the virtual timestamps are updated by the estimated execution time based on different clients' computational and communication capacities.
This enables us to simulate different response speeds and participating degrees of clients, which correspond to heterogeneous real-world mobile devices.
Moreover, in Sec.\ref{append:dp}, we show that pFL-Bench supports the exploration of trade-offs between pFL and privacy protection techniques, and conduct demonstrative experiments with Differential Privacy \cite{dp_survey}.

\section{Conclusions}
In this paper, we propose a comprehensive, standardized, and extensible benchmark for personalized Federated Learning (pFL), pFL-Bench, which contains 12 dataset variants with a wide range of domains and unified partitions, and more than 20 pFL methods with pluggable and easy-to-extend pFL subroutines.
We conduct extensive and systematic comparisons and conclude that designing effective, efficient and robust pFL methods with good generalization and convergence still remains challenging. 
We release pFL-Bench with guaranteed maintenance for the community, and believe that it will benefit reproducible, easy and generalizable pFL researches and potential applications.
We also welcome contributions of new pFL methods and datasets to keep pFL-Bench up-to-date.
\clearpage
\newpage

\bibliographystyle{unsrt}
\bibliography{pfl}

\clearpage

\newpage
\appendix

\section*{Appendices for the Paper: \textit{pFL-Bench: A Comprehensive Benchmark for Personalized Federated Learning}}

We provide more details and experimental results for pFL-Bench in the appendices:
\begin{itemize}
	\item Sec.\ref{append:data-detail}: the details of adopted datasets and models (\eg, tasks, heterogeneous partitions, and model architectures), and the extensions for other datasets and models with pFL-Bench.
	\item Sec.\ref{append:method-detail}: detailed description of methods and metrics in our experiments.
	\item Sec.\ref{append:implement-detail}: implementation details including the experimental environments and hyper-parameters.
	\item Sec.\ref{append:more-exp-results}: more experimental results in terms of generalization (Sec. \ref{append:general-results}), fairness (Sec. \ref{append:fair-results}) and efficiency (Sec.\ref{append:efficiency-results}) for all the datasets in Table 1. Besides, to demonstrate the potential and ease of extensibility of the pFL-bench, we also conducted experiments in the heterogeneous device resource scenario based on FedScale \cite{fedscale-icml} (Sec.\ref{append:hetero-device}), as well as experiments incorporating privacy-preserving techniques (Sec.\ref{append:dp}).
\end{itemize}

\section{Datasets and Models}
\label{append:data-detail}

\begin{figure}[t]
	\centering
	\subfigure{
		\begin{minipage}[]{\linewidth}
			\centering
			\includegraphics[width=0.99\linewidth]{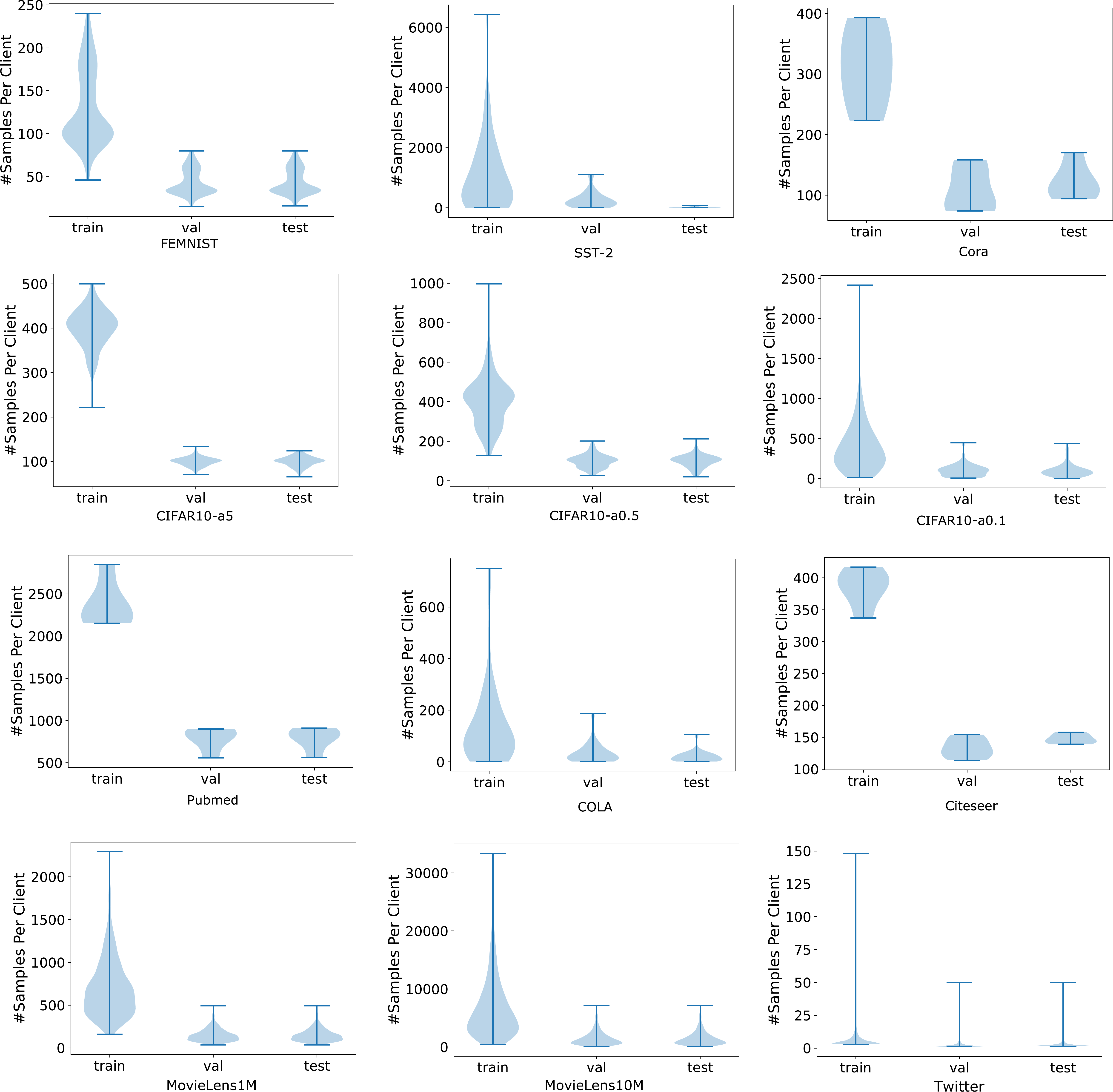}
			\caption{The violin plot of number of samples per client for all the adopted datasets. }
			\label{fig:client-statis-all}
		\end{minipage}%
	}
\end{figure}

\begin{figure}[t]
	\centering
	\subfigure{
		\begin{minipage}[]{\linewidth}
			\centering
			\includegraphics[width=0.99\linewidth]{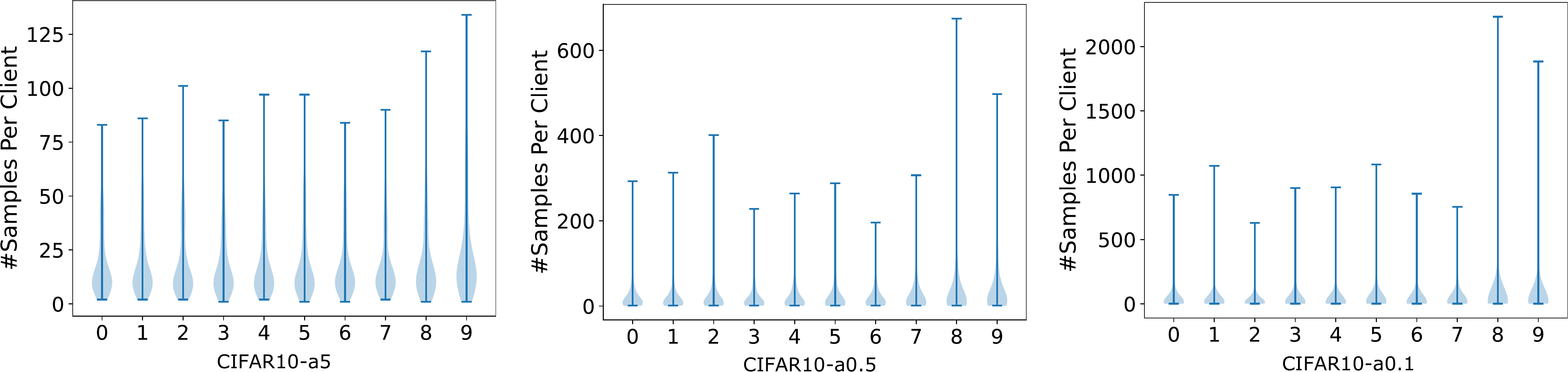}
			\caption{The label skew visualization in terms of number of labels per client for the CIFAR-10 datasets with different Dirichlet allocation factor $\alpha$s. }
			\label{fig:label-skew}
		\end{minipage}%
	}
\end{figure}

\begin{figure}[t]
	\centering
	\subfigure{
		\begin{minipage}[]{\linewidth}
			\centering
			\includegraphics[width=0.99\linewidth]{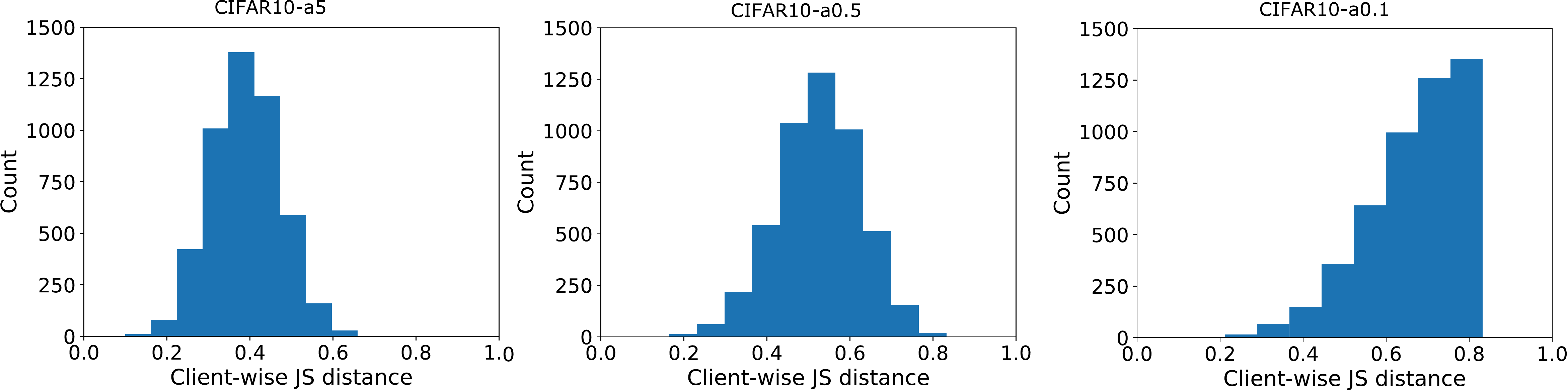}
			\caption{The histogram of clients' pairwise Jensen–Shannon distance in terms of their label distributions. The smaller the Jensen-Shannon distance, the more similar the compared distributions.}
			\label{fig:label-similar}
		\end{minipage}%
	}
\end{figure}

\paragraph{Experimental datasets.}
We present detailed descriptions of the 12 publicly available dataset variants  used in pFL-Bench.
These datasets are popular in the corresponding fields, and cover a wide range of domains, scales, partition manners, and Non-IID degrees. 
\begin{itemize}[leftmargin=*]
	\item The Federated Extended MNIST (\textbf{FEMNIST}) is a widely used FL dataset for 62-class handwritten character recognition \cite{caldas2018leaf}. The original FEMNIST dataset contains 3,550 clients and each client corresponds to a character writer from EMNIST \cite{cohen2017emnist}. Following \cite{xie2021federated}, we adopt the sub-sampled version in FL-Bench, which contains 200 clients and totally 43,400 images with resolution of 28x28 pixels, and the dataset is randomly split into train/valid/test sets with ratio 3:1:1. \footnote{For all the adopted datasets, the train/val/test splitting is conducted} within the local data of each client.
	In pFL-Bench, we use this dataset to vary the client sampling rate in FL processes as shown in Figure 4 in the main body of the paper. 
	\item The \textbf{CIFAR10} is a popular dataset for 10-class image classification containing 60,000 colored images with resolution of 32x32 pixels.
	Follow the heterogeneous partition manners used in \cite{marfoq2021fedEM,caldas2018leaf,diaoHeteroFLComputationCommunication2020,fallahPersonalizedFederatedLearning2020}, we use Dirichlet allocation to split this datasets into 100 clients with different Dirichlet factors as $\alpha=[5, 0.5, 0.1]$ (a smaller $\alpha$ indicates a higher heterogeneous degree).
	We split the dataset into train/valid/test sets with ratio 4:1:1.
	\item Corpus of Linguistic Acceptability (\textbf{COLA}) is a textual classification datasets from \cite{wang2018glue,warstadt2019neural}, which contains 9,600 English sentences labeled with grammatical correctness.
	In pFL-Bench, this dataset is partitioned into 50 clients via Dirichlet allocation with $\alpha=0.4$.
	We split the dataset into train/valid/test sets with a ratio of about 7:2:1.
	\item The Stanford Sentiment Treebank  (\textbf{SST-2}) is a sentiment classification dataset from \cite{wang2018glue,socher2013recursive}, which contains 68,200 movie reviews sentences labeled with human sentiment. Similar to COLA, in pFL-Bench, this dataset is partitioned into 50 clients with Dirichlet allocation and $\alpha=0.4$. The train/valid/test sets are with a ratio of about 60:15:1. For \textbf{COLA} and \textbf{SST-2}, since the test subsets from GLUE \cite{wang2018glue} are unlabeled (private in the GLUE server), we made new train/val/test partitions different from GLUE versions.
	\item The \textbf{Twitter} dataset is a textual sentiment analysis dataset from \cite{caldas2018leaf}. We adopt a subset which contains 13,203 users, the partition manner for this dataset is natural w.r.t. users, and the median number of data samples per user is 7. The train/valid/test sets for each client are with a ratio of about 3:1:1.
	\item The \textbf{Cora} dataset is a citation network that contains 2,708 nodes and 5,429 edges, in which each node indicates a scientific publication classified into one of seven classes \cite{mccallum2000automating}. Following FS-G \cite{wang2022federatedscopegnn}, we split it into 5 clients using a community detection algorithm, Louvain \cite{blondel2008fast}. The train/valid/test sets are with ratio about 3:1:1. 
	\item The \textbf{Pubmed} dataset contains 19,717 nodes and 44,338 edges.
	The nodes indicate scientific publications classified into one of three classes \cite{namata2012query}. Following FS-G \cite{wang2022federatedscopegnn}, we split it into 5 clients with Louvain community partition. The train/valid/test sets are with a ratio of about 3:1:5.
	\item The \textbf{Citeseer} dataset is a citation network that contains 3,312 nodes and 4,732 edges, in which each node indicates a scientific publication classified into one of six classes \cite{getoor2005link}. Following FS-G \cite{wang2022federatedscopegnn}, we split it into 5 clients with Louvain community partition. The train/valid/test sets are with a ratio of about 4:1:1.
	\item The \textbf{Movielens1M} contains 1,000,209 ratings from 6,040 users and 3,900 movies \cite{harper2015movielens}. Following the horizontal partition manner used in \cite{li2021federated}, in pFL-Bench, we split this dataset into 1,000 clients according to \textit{users}. The train/valid/test sets with ratio about 14:3:3.
	\item The \textbf{Movielens10M} contains 10,000,054 ratings from 71,567 users and 10,681 movies \cite{harper2015movielens}. Following the vertical partition manner used in \cite{li2021federated}, in pFL-Bench, we split this dataset into 1,000 clients according to \textit{items}. The train/valid/test sets are with ratio about 14:3:3.
\end{itemize}
For all these experimental datasets, we randomly select 20\% clients as new clients that do not participate in the FL processes.
We summarize some statistics in Table 1 in the main body of the paper.
Besides, we illustrate the violin plot of data size per client in Figure \ref{fig:client-statis-all},  the label skew visualization of certain datasets in Figure \ref{fig:label-skew}, and clients' pairwise similarity of label distribution in terms of Jensen–Shannon distance in Figure \ref{fig:label-similar}. And the smaller the Jensen-Shannon distance, the more similar the compared distributions.  We can see that as the degree of heterogeneity increase (the $\alpha$ decreases), the larger the label skew degree and the Jensen-Shannon distances we get. We can perform similar calculations on a variety of FL datasets, and further rank this distances, and in turn select those clients whose distributions are very different but whose models do not perform well for further analysis, understanding and algorithm improvement. 
Furthermore, all these results show diverse properties across the adopted FL datasets in pFL-Bench, enabling comprehensive comparisons among different methods.

\paragraph{Models}
To align with previous works \cite{reddi2021adaptive,marfoq2021fedEM,dinh2020personalized,liang2020think,shamsian2021personalized}, we preset a 2-layer CNN for FEMNIST and CIFAR10. 
Specifically, the model consists of two convolutional layers with 5 × 5 kernels, max pooling, batch normalization, ReLU activation, and two dense layers.
The hidden size is 2,048 and 512 for FEMNIST and CIFAR10 respectively.
For the COLA and SST-2 datasets, we preset the pre-trained BERT-Tiny model from \cite{turc2019}, which contains 2-layer Transformer encoders with a hidden size of 128.
For the Twitter dataset, we preset a LR model with 50d Glove embeddings \footnote{https://nlp.stanford.edu/data/glove.6B.zip}.
For the graph datasets, we preset the graph isomorphism neural network, GIN \cite{xu2018how}, which contains 2-layer convolutions with batch normalization, the hidden size of 64, and dropout rate of 0.5.
For the recommendation datasets, we preset the Matrix Factorization (MF) model  \cite{koren2009matrix} with a hidden size of 20 for user and item embeddings.

\paragraph{Remark on the adopted dataset scales and model sizes.}
It is worth noting that simulation with pFL algorithms on a large client scale is very challenging, due to the fact that we need to maintain distinct (personalized) model object for each client. Let's take the famous benchmark FEMNIST as an example, which has 3,550 users and suppose we adopt the widely-used two-layer CNN network. Although this model only occupies ~200MB, maintaining 3,550 such models would consume more than ~700GB memory. Different from non-personalized FL algorithms, for which it is feasible to maintain only one model object for all the clients, for pFL algorithms, we may have to switch and cache the personalized models among CPU, GPU and even disks. Due to the large number of methods and datasets included in our benchmark, and the corresponding huge hyper-parameter search space, we used several subsets of the FL datasets to reduce the reproduction and experimental barriers.

\paragraph{Extension.} 
We note that besides the experimental datasets and models introduced above, our code-base is compatible with a large number of datasets from other public popular DataZoos and ModelZoos.
We provide the unified dataset, dataloader, and model I/O interfaces with carefully designed modularity, which enables users to easily register and extend the datasets/models with simple and flexible configuration, such as different heterogeneous partition manners, number of clients, new client ratio, model types and model parameter dimensions.
Currently, we support datasets from LEAF~\cite{caldas2018leaf},  Torchvision \cite{torchvision}, Huggingface datasets \cite{lhoest-etal-2021-datasets}, FederatedScope (FS) \cite{xie2021federated} and FederatedScope-GNN (FS-G) \cite{wang2022federatedscopegnn}; and models from
Torchvision \cite{torchvision}, Huggingface \cite{wolf-etal-2020-transformers},  FS \cite{xie2021federated} and  FS-G \cite{wang2022federatedscopegnn}.

\section{Methods and Metrics}
\label{append:method-detail}
\subsection{Methods}
We present detailed descriptions of the methods in pFL-Bench, which conveys a range of popular and SOTA methods in three categories including \textbf{Non-pFL methods}, \textbf{pFL methods} and \textbf{Combined variants}.

The following \textbf{Non-pFL methods} are considered in pFL-Bench:
\begin{itemize}[leftmargin=*]
	\item The \emph{Global-Train} method refers to training only a centralized model from all data merged from all clients. 
	\item The \emph{Isolated} method indicates that each client trains its' client-specific model without FL communication. 
	The \emph{Global-Train} and \emph{Isolated} methods provide a good reference to examine the benefits of pFL processes. 
	For these two methods, we omit the un-participated clients.
	
	\item The \emph{FedProx} \cite{blocal} method leverages proximal term to encourage the updated models at clients not to differ too much from the global model.

	\item In addition, we include the classical \emph{FedAvg} \cite{mcmahan2017communication} that average gradients weighted by data size of clients in each FL round.
	
	\item The \emph{FedOpt} \cite{FedOPT2020Asad} algorithm is also considered, which generalizes FedAvg by introducing an optimizer for the FL server. We use the SGD as the server optimizer for FedOpt and search its learning rate.
\end{itemize}

\textbf{pFL methods.} 
We consider the following representative SOTA  methods: 

\begin{itemize}[leftmargin=*]
	\item \emph{FedBN} \cite{li2020fedbn} is a simple yet effective pFL method aiming to handle the feature shift Non-IID challenge. It locally maintains the clients' batch normalization parameters without FL communication and aggregation. In pFL-Bench, we 
	generalize FedBN into the Transformer model by filtering out the layer normalization parameters.
	
	\item The \emph{Ditto} \cite{li2021ditto} is a pFL method aiming to improve the fairness and robustness of FL.
	For each client, Ditto maintains the local personalized model and global model at the same time. 
	The global model is trained with the same produce in FedAvg and the local model is trained with a personalized regularization according to the global model parameters.
	
	\item The \emph{pFedMe} \cite{dinh2020personalized} is a meta-learning based method and also regularizes the local models according to the global model parameters. The authors propose to use Moreau envelops based regularization to reduce the complexity caused by Hessian matrix computation, which is required by some meta-learning based pFL methods such as Per-FedAvg \cite{fallah2020personalized}.
	
	\item The pFL-Bench also contains multi-model based pFL methods. 
	\item The \emph{HypCluster} \cite{mansour2020three} method proposes to split clients into clusters and learns different personalized models for different clusters. The cluster is determined by performance on validation sets. In our experiments, we set the number of clusters as 3 for a fair comparison with FedEM.

	The \emph{FedEM} method \cite{marfoq2021fedEM} assumes the local data distribution is a mixture of multiple underlying distributions. 
	It learns a mixture of multiple local models with Expectation-Maximization algorithm to deal with the data heterogeneity, and can be easily extended to several clustering based and multi-task learning based method pFL methods. 
	In our experiments, we use 3 internal models for FedEM according to the authors' default choice.

\end{itemize}

\textbf{Combined variants}. 
It is worth noting that we provide pluggable re-implementations of numerous existing methods in pFL-Bench.
This modularity enables users can pick different personalized objects and behaviors to form a new pFL variant. 
We combine \emph{FedBN}, \emph{FedOpt}, and \emph{Fine-tuning (FT)} with other compatible methods.
The \emph{FedBN} combination indicates to make the batch/layer normalization parameters personalized and locally maintained.
The \emph{FedOpt} combination indicates introducing the server optimizer into the FL processes.
The \emph{Fine-tuning (FT)} combination indicates fine-tuning the local models with a few steps before evaluation within the FL processes.

To facilitate fine-grained ablations and systematic pFL study, we finally compare more than 20 pFL method variants in the experiments. 
We will continuously include more pFL methods into pFL-Bench.

\subsection{Metrics}
Here we summarize the monitored metrics in our benchmark and give more detailed description about them.

\paragraph{Generalization.} We support server-side and clients-sides monitoring w.r.t. widely used performance metrics such as accuracy, loss, F1, etc. Specifically, in the paper, we denote $\overline{Acc}$ \ $\overline{Loss}$ be the accuracy \ Loss average weighted by the number of local data samples, $\widetilde{Acc}$ \ $\widetilde{Loss}$ be the accuracy \ Loss of un-participated clients, and $\Delta$ be the participation generalization gap.

\paragraph{Fairness.} We support different summarizing manners over the evaluated metrics over clients, such as weighted average (e.g., $\overline{Acc}$), uniform average (e.g., $\overline{Acc}'$), the standard deviation (denoted by $\sigma$ in our paper), and various quantiles such as bottom accuracy $\protect\widebreve{Acc}$. We report the $\lfloor |\mathcal{C}|/10 \rfloor$)-th worst accuracy where $|\mathcal{C}|$ is the number of all evaluated clients. To align with FedEM, the 90th percentile is considered here to omit the particularly noisy results from clients with worse performance with very small data sizes. 

\paragraph{System costs.} 
For computational and communication costs, we support to monitor some proxy metrics including FLOPs, communication bytes, and convergence rounds. The FLOPS are counted as the sum of amounts for both training and inference via a per-operator flops counting tool, fvcore/flop\_count. \footnote{https://github.com/facebookresearch/fvcore/blob/main/docs/flop\_count.md}
The reported communication bytes are counted as the sum of upstream and downstream across all participants until convergence with early stopping.
Besides, thanks to the good integration of wandb, our benchmark also supports more runtime metrics including the dynamic utilization of CPU, GPU, memory, disk, etc. \footnote{https://docs.wandb.ai/ref/app/features/system-metrics}

\section{Implementation} 
\label{append:implement-detail}
\paragraph{Enviroments.}
We implement pFL-Bench based on the FS \cite{xie2021federated} package and PyTorch.
The experiments are conducted on a cluster of 8 Tesla V100 and 64 NVIDIA GeForce GTX 1080 Ti GPUs, each machine with 380G memory and Xeon Platinum 8163 2.50GHz CPU containing 96 cores.
Our experiments are conducted in the containerized environments with Ubuntu18.04. 

We provide versioned \textit{DockerFiles}, the built docker images and experimental datasets in our website with Aliyun storage service. \footnote{ https://github.com/alibaba/FederatedScope/tree/master/benchmark/pFL-Bench}
The pFL-Bench and the underlying FS \cite{xie2021federated} package is continuously developed and maintained by Data Analytics and Intelligence Lab (DAIL) of DAMO Academy. 
We will actively fix potential issues, track updates and Github release.

\paragraph{Hyper-parameters.}
For fair comparisons, we first use wandb sweep \footnote{https://docs.wandb.ai/sweeps} with the hyper-parameter searching (HPO) algorithm, HyperBand \cite{li2017hyperband}, to find the best hyper-parameters for all the methods on all datasets.
The validation sets are used and we employ early stopping with a large number of total FL rounds $T$.
We set this hyper-parameter to make almost all methods converge within $T$ rounds.
Specifically, for FEMNIST, CIFAR10, Movielens-1M, Movielens-10M and Twitter, we set $T=1,000$.
For Cola, SST-2, Pubmed, Cora and Citeseer, we set $T=500$.
The batch size is set to be 32 for image datasets, 64 for textual datasets, and 1,024 for the recommendation datasets respectively.
For graph datasets, we adopt full batch training.
For all methods, we search the local update steps (\ie, the number of local training epochs in each FL round) from $[1, 3, 6]$.
For the local SGD learning rate, we search from $[0.05, 0.005, 0.5, 0.01, 0.1, 1, 2]$.
For FedOpt, we search the server learning rate from $[0.05, 0.1, 0.5, 1.5]$.
For pFedMe, we search the personalized regularization weight  from $[0.05, 0.1, 0.2, 0.9]$ and its local meta-learning step from $[1, 3]$.
For FedEM, we set its number of mixture models as 3.
For Ditto, we search its personalized regularization weight  from $[0.05, 0.1, 0.2, 0.5, 0.8]$.

To enable easily reproducible research, we provide standardized and documented scripts including the HPO scripts, the experiments running scripts and searched best configuration files in our code-base (see the link in the above paragraph).

\section{Additional Experimental Results}
\label{append:more-exp-results}
\subsection{Generalization}
\label{append:general-results}
The generalization results for FEMNIST, SST-2 and PUBMED are shown in the Table 2 and Figure 3 in the main body of the paper. Here we present the results for other datasets including all the textual datasets (Table \ref{tab:all_nlp}), all the graph datasets (Table \ref{tab:all_graph}), and all the recommendation datasets (Table \ref{tab:all_rec}). We note that FedOpt may gain bad results on some datasets when the models contain batch/layer normalization parameters. 
Besides, for the Twitter and recommendation datasets, we have not compared the FedBN based methods as the used LR and MF models do not contain batch/layer normalization parameters.

\begin{table}[]
	\centering
	
	\caption{Accuracy results for both participated clients and un-participated clients on COLA, SST-2 and Twitter datasets. $\overline{Acc}$ indicates the aggregated accuracy weighted by the number of local data samples of participated clients, $\widetilde{Acc}$ indicates the aggregated accuracy of un-participated clients, and $\Delta$ indicates the participation generalization gap. \textbf{Bold} and \underline{underlined} indicate the best and second-best results among all compared methods, while \textcolor{red}{red} and \textcolor{blue}{blue} indicate the best and second-best results for original methods without combination ``-''.
	}
	\resizebox{\columnwidth}{!}{
	\begin{tabular}{l|ccc|ccc|ccc}
		\toprule
		& \multicolumn{3}{c|}{COLA} & \multicolumn{3}{c}{SST-2} & \multicolumn{3}{c}{Twitter}\\
		& $\overline{Acc}$ & $\widetilde{Acc}$ & $\Delta$ & $\overline{Acc}$ & $\widetilde{Acc}$ & $\Delta$ & $\overline{Acc}$ & $\widetilde{Acc}$ & $\Delta$ \\
		\midrule
		Global-Train&69.06&-&-&\color{red}{\textbf{80.57}}&-&-&55.56&-&-\\
		Isolated&55.96&-&-&60.82&-&-&\color{blue}{70.04}&-&-\\
		FedAvg&\color{blue}{71.85}&\color{blue}{63.49}&-8.36&74.88&\color{red}{80.24}&\color{red}{5.36}&62.15&61.24&\color{blue}{-0.91}\\
		FedAvg-FT&68.29&58.66&-9.63&74.14&\textbf{83.28}&9.13&70.53&71.17&\textbf{0.64}\\
		FedOpt&71.85&59.62&-12.23&72.28&\underline{83.06}&10.78&62.09&61.64&-0.45\\
		FedOpt-FT&62.59&47.82&-14.77&65.77&80.02&14.25&\underline{71.08}&71.41&0.33\\
		\midrule
		pFedMe&\color{red}{\underline{74.40}}&\color{red}{\underline{67.64}}&\color{blue}{-6.76}&71.27&69.34&-1.92&63.45&\color{blue}{62.52}&-0.94\\
		pFedMe-FT&\textbf{78.47}&\textbf{76.33}&-2.14&75.61&66.48&-9.13&\textbf{84.00}&\textbf{71.80}&-12.20\\
		\midrule
		FedBN&\color{blue}{71.85}&\color{blue}{63.49}&-8.36&74.88&\color{blue}{75.40}&\color{blue}{0.52}&-&-&-\\
		FedBN-FT&66.71&49.87&-16.84&68.81&82.43&13.63&-&-&-\\
		FedBN-FedOPT&71.85&62.48&-9.37&64.70&65.50&0.81&-&-&-\\
		FedBN-FedOPT-FT&67.48&57.59&-9.90&68.65&70.56&1.91&-&-&-\\
		\midrule
		Ditto&55.46&49.90&\color{red}{-5.56}&52.03&46.79&-5.24&\color{red}{70.23}&49.60&-20.63\\
		Ditto-FT&72.11&52.15&-19.96&56.49&65.50&9.01&69.99&51.32&-18.67\\
		Ditto-FedBN&70.69&49.90&-20.79&56.03&46.79&-9.24&-&-&-\\
		Ditto-FedBN-FT&72.66&53.44&-19.21&53.15&66.49&13.34&-&-&-\\
		Ditto-FedBN-FedOpt&50.25&49.90&-0.35&57.67&46.79&-10.88&-&-&-\\
		Ditto-FedBN-FedOpt-FT&55.01&58.22&\textbf{3.21}&52.89&66.49&13.60&-&-&-\\
		\midrule
		FedEM&\color{blue}{71.85}&\color{blue}{63.49}&-8.36&\color{blue}{\underline{75.78}}&67.67&-8.11&63.44&\color{red}{62.68}&\color{red}{-0.75}\\
		FedEM-FT&54.90&48.29&-6.61&64.86&81.63&\textbf{16.77}&70.97&\underline{71.59}&\underline{0.62}\\
		FedEM-FedBN&71.44&63.99&-7.45&75.43&62.81&-12.62&-&-&-\\
		FedEM-FedBN-FT&57.62&58.88&1.26&64.96&81.04&\underline{16.08}&-&-&-\\
		FedEM-FedBN-FedOPT&71.85&62.82&-9.03&72.25&64.69&-7.56&-&-&-\\
		FedEM-FedBN-FedOPT-FT&57.23&58.88&\underline{1.65}&62.26&73.87&11.61&-&-&-\\
		\bottomrule
	\end{tabular}
	}
	\label{tab:all_nlp}
\end{table}

\begin{table}[!t]
	\centering
	
	\caption{Accuracy results for both participated clients and un-participated clients on Pubmed, Cora and Citeseer datasets. $\overline{Acc}$ indicates the aggregated accuracy weighted by the number of local data samples of participated clients, $\widetilde{Acc}$ indicates the aggregated accuracy of un-participated clients, and $\Delta$ indicates the participation generalization gap. \textbf{Bold} and \underline{underlined} indicate the best and second-best results among all compared methods, while \textcolor{red}{red} and \textcolor{blue}{blue} indicate the best and second-best results for original methods without combination ``-''.
	}
	\vspace{-0.05in}
	\small
	\resizebox{\columnwidth}{!}{
		\begin{tabular}{p{1.42in}|ccc|ccc|ccc}
			\toprule
			& \multicolumn{3}{c|}{PUBMED} & \multicolumn{3}{c|}{CORA} & \multicolumn{3}{c}{CITESEER}  \\
			& $\overline{Acc}$ & $\widetilde{Acc}$ & $\Delta$ & $\overline{Acc}$ & $\widetilde{Acc}$ & $\Delta$& $\overline{Acc}$ & $\widetilde{Acc}$ & $\Delta$ \\
			\midrule
			Global-Train&87.01&-&-&\color{red}{\textbf{86.10}}&-&-&74.03&-&-\\
			Isolated&85.56&-&-&82.48&-&-&69.83&-&-\\
			FedAvg&\color{blue}{87.27}&\color{red}{72.63}&\color{blue}{-14.64}&81.30&\color{red}{72.14}&\color{red}{-9.16}&\color{blue}{75.58}&\color{red}{59.83}&\color{red}{-15.74}\\
			FedAvg-FT&87.21&\underline{79.78}&\underline{-7.43}&82.07&75.84&\underline{-6.22}&75.63&66.07&-9.57\\
			FedOpt&67.38&53.84&-13.54&70.70&59.58&-11.12&71.59&55.16&-16.43\\
			FedOpt-FT&82.36&64.09&-18.27&82.68&62.17&-20.51&74.34&62.36&-11.99\\
			\midrule
			pFedMe&86.91&\color{blue}{71.64}&-15.27&83.18&70.13&-13.05&75.30&58.45&-16.85\\
			pFedMe-FT&85.71&77.07&-8.64&82.11&71.48&-10.63&75.35&62.14&-13.20\\
			\midrule
			FedBN&\color{red}{\textbf{88.49}}&52.53&-35.95&\color{blue}{84.13}&57.33&-26.80&\color{red}{75.80}&51.29&-24.51\\
			FedBN-FT&87.45&\textbf{80.36}&\textbf{-7.09}&76.20&64.13&-12.07&75.07&64.20&-10.87\\
			FedBN-FedOPT&87.87&42.72&-45.15&84.64&53.27&-31.37&76.20&50.34&-25.86\\
			FedBN-FedOPT-FT&87.54&77.07&-10.47&84.10&68.14&-15.96&\textbf{76.70}&62.84&-13.86\\
			\midrule
			Ditto&\color{blue}{87.27}&2.84&-84.43&83.67&14.53&-69.14&74.79&14.52&-60.27\\
			Ditto-FT&87.47&35.03&-52.44&81.47&72.64&-8.84&76.47&39.27&-37.20\\
			Ditto-FedBN&\underline{88.18}&2.84&-85.34&81.38&14.53&-66.84&75.35&14.52&-60.83\\
			Ditto-FedBN-FT&87.83&28.52&-59.30&83.25&65.87&-17.38&75.97&36.51&-39.46\\
			Ditto-FedBN-FedOpt&87.81&2.84&-84.97&82.54&14.53&-68.01&75.07&14.52&-60.55\\
			Ditto-FedBN-FedOpt-FT&87.60&18.18&-69.42&82.00&70.75&-11.25&76.42&48.99&-27.43\\
			\midrule
			FedEM&85.64&71.12&\color{red}{-14.52}&81.92&\color{blue}{72.02}&\color{blue}{-9.90}&75.41&\color{blue}{59.60}&\color{blue}{-15.81}\\
			FedEM-FT&85.88&78.08&-7.80&77.32&\textbf{79.45}&\textbf{2.13}&72.71&66.77&\textbf{-5.94}\\
			FedEM-FedBN&88.12&48.64&-39.48&\underline{85.07}&50.98&-34.09&74.90&41.55&-33.35\\
			FedEM-FedBN-FT&86.38&72.02&-14.35&84.61&76.77&-7.83&75.29&\underline{68.38}&\underline{-6.91}\\
			FedEM-FedBN-FedOPT&87.56&42.37&-45.19&84.68&56.45&-28.24&76.08&53.81&-22.27\\
			FedEM-FedBN-FedOPT-FT&87.49&72.39&-15.09&85.02&\underline{76.97}&-8.05&\underline{76.59}&\textbf{68.62}&-7.96\\
			\bottomrule
	\end{tabular}}
	\label{tab:all_graph}
\end{table}

\begin{table}[!t]
	\centering
	\caption{Accuracy results for both participated clients and un-participated clients on Movielens-1M and Movielens-10M datasets. $\overline{Loss}$ indicates the loss average weighted by the number of local data samples, $\widetilde{Loss}$ indicates the loss of un-participated clients, and $\Delta$ indicates the participation generalization gap. \textbf{Bold} and \underline{underlined} indicate the best and second-best results among all compared methods, while \textcolor{red}{red} and \textcolor{blue}{blue} indicate the best and second-best results for original methods without combination ``-''.
	}
	\begin{tabular}{p{1.42in}|ccc|ccc}
		\toprule
		& \multicolumn{3}{c|}{Movielens-1M} & \multicolumn{3}{c}{Movielens-10M} \\
		& $\overline{Loss}$ & $\widetilde{Loss}$ & $\Delta$ & $\overline{Loss}$ & $\widetilde{Loss}$ & $\Delta$ \\
		\midrule
		Global-Train&\color{blue}{0.78}&-&-&\color{red}{\textbf{0.67}}&-&-\\
		Isolated&10.35&-&-&11.48&-&-\\
		FedAvg&0.84&\color{red}{14.17}&\color{blue}{13.33}&\color{blue}{\underline{0.70}}&\color{blue}{13.39}&12.68\\
		FedAvg-FT&0.84&9.76&8.92&0.71&\textbf{11.07}&\underline{10.36}\\
		FedAvg-FT-FedOpt&0.85&5.15&4.31&0.73&\underline{11.40}&10.66\\
		FedOpt&0.83&14.17&13.34&0.71&13.39&12.68\\
		FedOpt-FT&0.83&12.06&11.23&0.74&11.92&11.18\\
		\midrule
		pFedMe&\color{red}{\textbf{0.54}}&\color{blue}{14.18}&13.64&13.06&\color{red}{12.73}&\color{red}{\textbf{-0.33}}\\
		pFedMe-FT&\underline{0.60}&8.20&7.60&0.80&12.59&11.79\\
		\midrule
		Ditto&1.29&14.19&\color{red}{12.89}&1.84&\color{blue}{13.39}&\color{blue}{11.55}\\
		Ditto-FT&1.35&14.17&12.81&1.69&13.39&11.70\\
		Ditto-FT-FedOpt&1.36&14.15&12.79&2.03&13.39&11.35\\
		\midrule
		FedEM&0.85&14.27&13.43&1.75&13.41&11.65\\
		FedEM-FT&0.85&\underline{4.86}&\underline{4.01}&0.87&12.80&11.93\\
		FedEM-FT-FedOpt&0.86&\textbf{4.47}&\textbf{3.61}&1.43&13.25&11.82\\
		\bottomrule
	\end{tabular}
	\label{tab:all_rec}
\end{table}

\subsection{Fairness}
\label{append:fair-results}
The fairness results for FEMNIST, SST-2 and PUBMED are listed in Table 3 in the main body of the paper. 
Here we present the results for other datasets, including all the textual datasets (Table \ref{tab:all_nlp_fair}), all the graph datasets (Table \ref{tab:all_graph_fair}), and all the recommendation datasets (Table \ref{tab:all_rec_fair}).

\begin{table}[!h]
	\centering
	\caption{Fairness results on COLA, SST-2 and Twitter datasets. $\overline{Acc}'$ indicate the equally-weighted average, $\sigma$ indicating the standard deviation of the average accuracy, and $\protect\widebreve{Acc}$ indicating the bottom accuracy. \textbf{Bold}, \underline{underlined}, \textcolor{red}{red} and \textcolor{blue}{blue} indicate the same highlights as used in Table \ref{tab:general}.}
	\resizebox{\columnwidth}{!}{
	\begin{tabular}{l|ccc|ccc|ccc}
		\toprule
		& \multicolumn{3}{c|}{COLA} & \multicolumn{3}{c}{SST-2} & \multicolumn{3}{c}{Twitter} \\
		& $\overline{Acc}'$ & $\sigma$ & $\protect\widebreve{Acc}$ &  $\overline{Acc}'$ & $\sigma$ & $\protect\widebreve{Acc}$ &  $\overline{Acc}'$ & $\sigma$ & $\protect\widebreve{Acc}$ \\
		\midrule
		Isolated&56.86&\color{red}{35.34}&\color{blue}{0.00}&59.40&41.29&0.00&\color{red}{67.44}&37.86&\color{red}{\underline{0.00}}\\
		FedAvg&51.53&\color{blue}{35.96}&\color{blue}{0.00}&\color{blue}{\underline{76.30}}&\color{red}{\underline{22.02}}&\color{red}{\underline{44.85}}&56.98&37.97&\color{red}{\underline{0.00}}\\
		FedAvg-FT&58.74&35.21&0.00&75.36&27.67&31.08&68.45&\underline{36.19}&\underline{0.00}\\
		FedOpt&57.10&\textbf{31.70}&\textbf{10.85}&73.78&34.03&17.53&59.95&37.88&\underline{0.00}\\
		FedOpt-FT&59.77&35.99&0.00&66.17&33.73&15.56&69.20&36.28&\underline{0.00}\\
		\midrule
		pFedMe&\color{red}{\underline{67.58}}&38.06&\color{red}{0.90}&65.08&26.59&27.75&58.18&\color{red}{36.67}&\color{red}{\underline{0.00}}\\
		pFedMe-FT&\textbf{69.17}&\underline{33.63}&\underline{9.90}&74.36&27.02&32.49&\textbf{78.82}&\textbf{33.28}&\textbf{22.22}\\
		\midrule
		FedBN&\color{blue}{59.60}&\color{blue}{35.96}&\color{blue}{0.00}&\color{blue}{\underline{76.30}}&\color{red}{\underline{22.02}}&\color{red}{\underline{44.85}}&-&-&-\\
		FedBN-FT&59.69&35.44&0.00&68.50&26.83&29.17&-&-&-\\
		FedBN-FedOPT&59.60&36.03&0.00&65.59&31.07&22.22&-&-&-\\
		FedBN-FedOPT-FT&59.10&35.15&0.00&68.42&28.18&30.71&-&-&-\\
		\midrule
		Ditto&55.14&36.76&\color{blue}{0.00}&49.94&40.81&0.00&\color{blue}{66.90}&38.05&\color{red}{\underline{0.00}}\\
		Ditto-FT&63.61&35.02&0.00&54.34&39.26&0.00&66.91&38.08&\underline{0.00}\\
		Ditto-FedBN&62.68&35.74&0.00&49.44&41.80&0.00&-&-&-\\
		Ditto-FedBN-FT&63.58&34.58&0.00&52.18&39.85&0.00&-&-&-\\
		Ditto-FedBN-FedOpt&52.48&35.89&0.00&55.61&40.43&1.39&-&-&-\\
		Ditto-FedBN-FedOpt-FT&57.13&36.20&0.00&53.16&34.75&9.72&-&-&-\\
		\midrule
		FedEM&51.52&\color{blue}{35.96}&\color{blue}{0.00}&\color{red}{\textbf{76.53}}&\color{blue}{23.34}&\color{blue}{44.44}&61.70&\color{blue}{37.72}&\color{red}{\underline{0.00}}\\
		FedEM-FT&57.80&35.24&0.00&64.29&32.84&12.96&\underline{70.19}&36.35&\underline{0.00}\\
		FedEM-FedBN&57.95&34.11&1.52&75.06&\textbf{18.48}&\textbf{53.33}&-&-&-\\
		FedEM-FedBN-FT&58.74&35.56&1.00&64.33&35.72&8.59&-&-&-\\
		FedEM-FedBN-FedOPT&59.60&35.49&0.00&72.66&27.18&34.17&-&-&-\\
		FedEM-FedBN-FedOPT-FT&56.74&35.61&1.00&58.42&31.21&17.93&-&-&-\\
		\bottomrule
	\end{tabular}
	}
	\label{tab:all_nlp_fair}
\end{table}

\begin{table}[!h]
	\centering
	\small
	\caption{Fairness results on Pubmed, Cora, and Citeseer datasets. $\overline{Acc}'$ indicate the equally-weighted average, $\sigma$ indicating the standard deviation of the average accuracy, and $\protect\widebreve{Acc}$ indicating the bottom accuracy. \textbf{Bold}, \underline{underlined}, \textcolor{red}{red} and \textcolor{blue}{blue} indicate the same highlights as used in Table \ref{tab:general}.}
	\resizebox{\columnwidth}{!}{
		\begin{tabular}{l|ccc|ccc|ccc}
			\toprule
			& \multicolumn{3}{c|}{PUBMED} & \multicolumn{3}{c|}{CORA} & \multicolumn{3}{c}{CITESEER} \\
			& $\overline{Acc}'$ & $\sigma$ & $\protect\widebreve{Acc}$ &  $\overline{Acc}'$ & $\sigma$ & $\protect\widebreve{Acc}$ & $\overline{Acc}'$ & $\sigma$ & $\protect\widebreve{Acc}$ \\
			\midrule
			Isolated&84.67&6.26&74.63&81.62&4.67&72.64&69.90&5.76&61.10\\
			FedAvg&86.72&\color{blue}{3.93}&79.76&81.07&5.06&73.26&\color{red}{75.64}&5.03&67.30\\
			FedAvg-FT&86.71&3.86&80.57&81.90&3.06&77.57&75.77&5.00&68.68\\
			FedOpt&66.69&16.69&48.50&70.17&12.56&38.89&71.60&9.20&42.25\\
			FedOpt-FT&81.53&16.69&46.21&82.31&6.35&68.03&74.38&4.13&69.26\\
			\midrule
			pFedMe&86.35&4.43&78.76&82.76&\color{red}{3.40}&\color{blue}{76.70}&75.36&4.79&\color{blue}{68.55}\\
			pFedMe-FT&85.47&\textbf{3.06}&80.95&81.98&\textbf{1.63}&79.58&75.40&4.39&70.31\\
			\midrule
			FedBN&\color{red}{\textbf{87.97}}&\color{red}{\underline{3.42}}&\color{red}{81.77}&\color{red}{83.64}&3.88&\color{red}{77.53}&\color{blue}{75.59}&\color{red}{\textbf{3.80}}&\color{red}{71.24}\\
			FedBN-FT&87.02&3.47&80.13&76.01&4.32&69.92&75.16&5.04&67.91\\
			FedBN-FedOPT&87.43&4.64&80.81&84.11&3.93&75.44&76.33&4.28&\textbf{72.43}\\
			FedBN-FedOPT-FT&87.02&3.94&81.78&83.79&2.39&\textbf{80.26}&\textbf{76.77}&4.60&71.68\\
			\midrule
			Ditto&\color{blue}{86.85}&3.98&\color{blue}{80.44}&\color{blue}{83.50}&\color{blue}{3.54}&75.02&75.55&\color{blue}{4.23}&67.99\\
			Ditto-FT&87.10&3.52&80.46&81.53&3.67&75.63&76.57&4.24&70.31\\
			Ditto-FedBN&\underline{87.75}&3.70&81.82&81.36&3.34&74.93&75.46&4.83&68.57\\
			Ditto-FedBN-FT&87.43&3.77&81.15&82.21&4.37&72.56&76.05&4.21&69.16\\
			Ditto-FedBN-FedOpt&87.27&3.90&79.14&82.38&\underline{2.10}&77.69&75.19&4.06&68.60\\
			Ditto-FedBN-FedOpt-FT&87.10&3.79&80.93&81.99&3.67&73.29&76.52&4.43&71.31\\
			\midrule
			FedEM&85.05&4.44&78.51&81.72&3.72&74.95&75.49&4.66&68.50\\
			FedEM-FT&85.54&4.48&79.39&78.43&11.72&56.20&72.88&6.55&63.53\\
			FedEM-FedBN&87.63&4.14&\textbf{82.54}&\underline{84.45}&3.14&76.71&74.29&4.22&69.06\\
			FedEM-FedBN-FT&85.68&4.33&79.44&\textbf{84.57}&3.04&\underline{80.22}&75.40&\underline{3.84}&70.59\\
			FedEM-FedBN-FedOPT&87.11&4.24&80.32&83.88&4.37&78.93&76.17&4.76&70.54\\
			FedEM-FedBN-FedOPT-FT&87.16&3.66&\underline{82.20}&84.40&4.43&79.53&\underline{76.74}&4.53&\underline{71.92}\\
			\bottomrule
	\end{tabular}}
	\label{tab:all_graph_fair}
\end{table}

\begin{table}[!h]
	\centering
	\caption{Fairness results on Movielens-1M and Movielens-10M datasets. $\overline{Loss}'$ indicate the equally-weighted average loss, $\protect\widebreve{Loss}$ indicates the bottom loss (the largest), and $\sigma$ indicates the standard deviation of the average loss. \textbf{Bold}, \underline{underlined}, \textcolor{red}{red} and \textcolor{blue}{blue} indicate the same highlights as used in Table \ref{tab:general}.}
	\begin{tabular}{l|ccc|ccc}
		\toprule
		& \multicolumn{3}{c|}{Movielens-1M} & \multicolumn{3}{c}{Movielens-10M}  \\
		& $\overline{Loss}'$ & $\sigma$ & $\protect\widebreve{Loss}$ &  $\overline{Loss}'$ & $\sigma$ & $\protect\widebreve{Loss}$ \\
		\midrule
		Isolated&11.12&2.43&14.34&11.44&1.83&13.75\\
		FedAvg&\color{blue}{0.85}&\color{blue}{0.21}&\color{blue}{1.13}&\color{red}{\textbf{0.71}}&\color{red}{\textbf{0.11}}&\color{red}{\textbf{0.84}}\\
		FedAvg-FT&0.85&0.21&1.12&\textbf{0.71}&\textbf{0.11}&\underline{0.85}\\
		FedAvg-FT-FedOpt&0.86&0.22&1.14&0.75&\underline{0.12}&0.90\\
		FedOpt&0.84&0.21&1.11&\underline{0.72}&\underline{0.12}&0.87\\
		FedOpt-FT&0.84&0.21&1.11&0.77&0.13&0.94\\
		\midrule
		pFedMe&\color{red}{\textbf{0.55}}&\color{red}{\textbf{0.11}}&\color{red}{\textbf{0.69}}&12.48&2.44&15.76\\
		pFedMe-FT&\underline{0.60}&\underline{0.12}&\underline{0.75}&0.80&0.13&0.96\\
		\midrule
		Ditto&1.31&0.77&1.69&\color{blue}{1.81}&\color{blue}{0.24}&\color{blue}{2.12}\\
		Ditto-FT&1.35&0.88&1.70&2.30&1.15&4.05\\
		Ditto-FT-FedOpt&1.35&0.79&1.70&1.98&0.27&2.32\\
		\midrule
		FedEM&0.87&0.22&1.15&2.37&1.25&4.09\\
		FedEM-FT&0.87&0.23&1.16&0.98&0.26&1.29\\
		FedEM-FT-FedOpt&0.87&0.22&1.16&1.88&0.94&3.09\\			
		\bottomrule
	\end{tabular}
	\label{tab:all_rec_fair}
\end{table}

\subsection{Efficiency}
\label{append:efficiency-results}
The efficiency-accuracy trade-off results for FEMNIST are plotted in Figure 5 in the main body of the paper. 
Here we present more efficiency-accuracy trade-off results for the experimental datasets, including the FEMNIST  datasets with different client sampling rates (Table \ref{tab:compute_femnist} and Table \ref{tab:converge_femnist}), 
CIFAR-10 datasets with different $\alpha$ (Table \ref{tab:compute_cifar} and Table \ref{tab:converge_cifar}), 
all the textual datasets (Table \ref{tab:compute_nlp} and Table \ref{tab:converge_nlp}), 
all the textual datasets (Table \ref{tab:compute_graph} and Table \ref{tab:converge_graph}), 
and all the recommendation datasets (Table \ref{tab:compute_rec} and Table \ref{tab:converge_rec}).

Besides the reported proxy system metrics such as FLOPs and the number of convergence rounds of FL processes,
our benchmark also supports monitoring more runtime metrics. Thanks to the good integration with wandb, we can easily track the usage of system resources in runtime including utilization of CPU, GPU, memory, disk, etc.
In Table \ref{tab:memory-time}, we report the average and peak process memory usage (in MB) and process running times (in seconds). In general, most pFL algorithms do have higher time and space overheads.
We omit to report results for other metrics since we started very many sets of experiments concurrently, taking up as much of the graphics card's memory and maximising CPU/GPU utilisation as possible, these metrics do not differ much from one of our different experiments. However, it is worth noting that these omitted metrics can be used to analyse algorithm bottlenecks in terms of system performance, and to optimise the space-time efficiency in single-experiment scenarios.
 
\subsection{Heterogeneous Device Resources}
\label{append:hetero-device}
The proposed pFL-Bench has good extensibility to support experiments in heterogeneous device resource scenarios, where clients have different computational and communication capacities.
Specifically, we integrate FedScale \cite{fedscale-icml} into our benchmark with a simulator that executes the behaviors of clients according to virtual timestamps of their message delivery to the server.
The virtual timestamps are updated by the estimated execution time based on clients' computational and communication capacities with the cost model proposed in FedScale.
The server employs an over-selection mechanism for clients at each broadcast round and thus some clients' message may be dropped, since the clients have different system capacities and different respond speeds corresponding to real-world mobile devices. \footnote{https://github.com/SymbioticLab/FedScale/tree/master/benchmark/dataset/data/device\_info}

Here we take the Ditto method on FEMNIST dataset as an example and present the results of experiments with heterogeneous device resources in Table \ref{tab:hetero-device}. Let $s$ to be the clients sampling rate for each FL round, and $s_{agg}$ be the the minimal ratio of received feedback w.r.t. the number of clients for the server to trigger federated aggregation in over-selection mode.
For the homo-device case, we set $s=0.2$ and for the hetero-device case, we set the $s=0.25$ and $s_{agg}=0.8$, leading to the same number of clients used for each federated aggregation.
From the results, we can see that the hetero-device version has slower convergence speed ($T'=0$ indicates that the early-stopping is not triggered within the large number of FL rounds $T=1000$), and it gains worse performance than the homo-device version, especially for the bottom accuracy ($\overline{Acc}'$) and standard deviation of the average accuracy ($\sigma$).
This shows unfairness among clients due to the fact that some low-resourced clients have too long computation or communication time to make their feedback incorporated into the federated aggregation, calling for more considerations w.r.t. device heterogeneity within pFL algorithms design.

\begin{table}[h]
	\centering
	
	\caption{Comparison between Ditto methods with and without heterogeneous device capabilities on FEMNIST dataset. $\overline{Acc}$ indicates the accuracy average weighted by the number of local data samples, $\widetilde{Acc}$ indicates the accuracy of un-participated clients, and $\Delta$ indicates the participation generalization gap. $\overline{Acc}'$ indicate the equally-weighted average, $\sigma$ indicating the standard deviation of the average accuracy, and $\protect\widebreve{Acc}$ indicating the bottom accuracy. Efficiency metrics include total FLOPS, communication bytes (Com.) and the convergence round $T'=0$. The $T'=0$ indicates the early-stopping is not triggered within the large number of FL rounds $T=1000$.
	}
	\vspace{-0.05in}
	\small
	\resizebox{\columnwidth}{!}{
		\begin{tabular}{c|ccc|ccc|ccc}
			\toprule
			& $\overline{Acc}$ & $\widetilde{Acc}$ & $\Delta$ & $\overline{Acc}'$ & $\sigma$ & $\protect\widebreve{Acc}$ & $T'$  & FLOPS & Com.\\
			\midrule
Ditto, Homo-Device &	88.39&	2.2	&-86.19&	87.18&	7.52	&78.23&	849.3G&	2.81M	&610 \\
Ditto, Hetero-Device&	79.76&	1.43&	-78.33&	77.39&	11.25&	61.76&	1.72T&	5.72M&	0 \\
			\bottomrule
	\end{tabular}}
	\label{tab:hetero-device}
\end{table}

\subsection{Incorporating Differential Privacy}
\label{append:dp}
It is interesting and under-explored to investigate the trade-off between personalization and privacy protection, which is important as FL involves the transmitting of local (maybe private) information.
We note that FederatedScope supports various privacy-related fundamental components and algorithms, such as Differential Privacy (DP) \cite{dwork2014dp} and privacy attack methods \cite{lyu2020attack} that can be used to examine the privacy-preserving strength.
Further research that combines pFL and privacy-preserving techniques will be convenient based on the modularized and extensible design of our benchmark. 
As a preliminary example, here we demonstrate the combination of the pFL with a Differential Privacy algorithm, the NbAFL \cite{wei2020nbafl} that achieves $(\epsilon, \delta)$-DP via noise injection and gradient clipping.

\begin{figure}[h]
	\centering
	\subfigure{
		\begin{minipage}[]{\linewidth}
			\centering
			\includegraphics[width=0.99\linewidth]{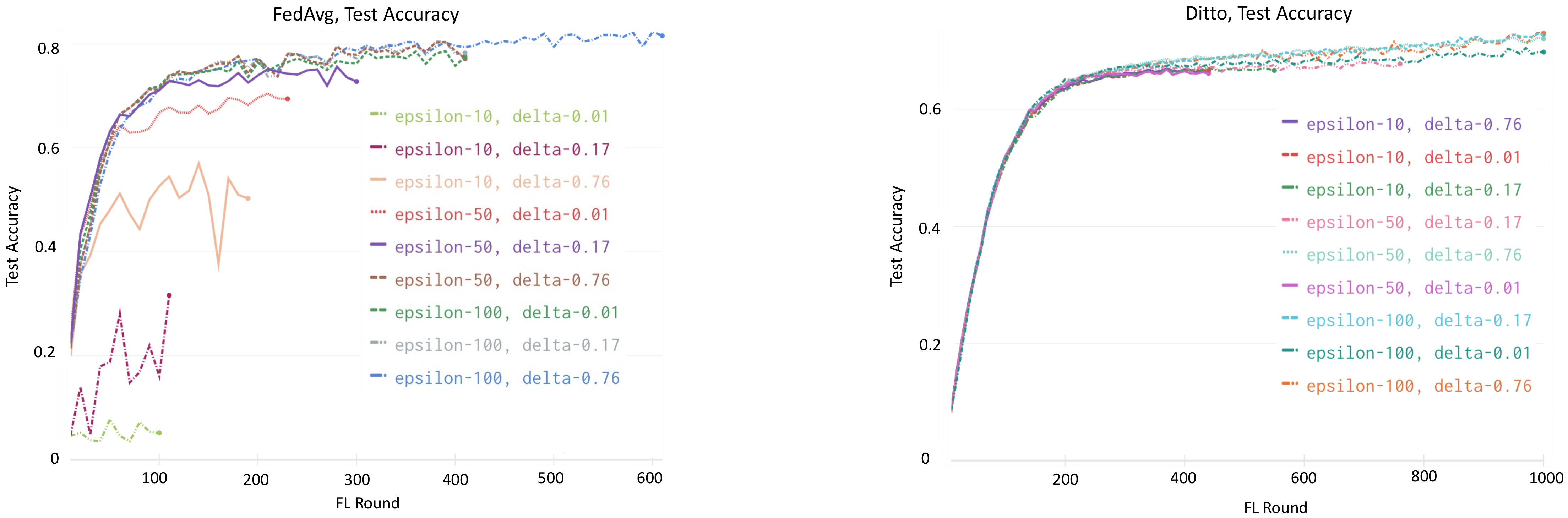}
			\caption{The learning curves of FedAvg and Ditto on FEMNIST dataset with various ($\epsilon, \delta$)-DPs. }
			\label{fig:dp-pfl}
		\end{minipage}%
	}
\end{figure}

In Figure \ref{fig:dp-pfl}, we plot the learning curves of FedAvg and Ditto methods on FEMNIST dataset with various ($\epsilon, \delta$)-DP.
Generally speaking, for privacy protection, the smaller the protection, the less performance degradation there is.
We can see that in the Figure, with larger $\epsilon$ and $\delta$, the accuracy ($\overline{Acc}$) is better for both the compared methods, which meets our expectation.
Interestingly, Ditto shows significantly better robustness for the dramatically varying privacy protection strengths during the whole learning process than FedAvg.
This may be because, in the noise perturbation scenarios, the personalized local model potentially brings up more local optimal points that can be reached for clients.
But there is still a gap for the best achievable performance between Ditto and FedAvg, leaving an interesting open question about how to reduce the performance degradation by co-designing personalization and noise injection.

\begin{sidewaystable}
	\centering
	\caption{The efficiency-accuracy trade-off results including total FLOPS, communication bytes (Com.), and $\overline{Acc}$ for FEMNIST datasets with different $s$.}
	\begin{tabular}{l|lll|lll|lll}
		\toprule
		& \multicolumn{3}{c|}{FEMNIST, $s=0.2$} & \multicolumn{3}{c|}{FEMNIST, $s=0.1$} & \multicolumn{3}{c}{FEMNIST, $s=0.05$} \\
		& FLOPS & Com. &  $\overline{Acc}$&  
		FLOPS & Com. &  $\overline{Acc}$&
		FLOPS & Com. &  $\overline{Acc}$ \\
		\midrule
		FedAvg&\color{red}{195.38G}&\color{blue}{1.24M}&83.97&\color{blue}{159.9G}&\color{red}{\underline{373.57K}}&83.84&\color{red}{\underline{78.48G}}&\color{red}{\underline{384.0K}}&83.21\\
		FedAvg-FT&296.8G&1.35M&86.44&155.49G&630.44K&85.45&217.97G&522.82K&84.78\\
		FedOpt&846.22G&1.68M&19.31&272.3G&635.48K&27.29&\textbf{29.0G}&\textbf{80.81K}&22.53\\
		FedOpt-FT&\textbf{77.81G}&\textbf{149.97K}&9.69&\textbf{54.88G}&\textbf{222.47K}&31.02&365.54G&877.15K&29.73\\
		\midrule
		pFedMe&2.24T&1.52M&\color{blue}{87.50}&516.47G&1012.93K&84.70&216.66G&\color{blue}{493.06K}&\color{blue}{85.23}\\
		pFedMe-FT&1.33T&2.19M&88.19&1.22T&953.21K&86.28&585.23G&569.72K&87.37\\
		\midrule
		FedBN&\color{blue}{243.31G}&\color{red}{1.21M}&86.72&\color{red}{\underline{120.66G}}&\color{blue}{510.15K}&\color{blue}{85.10}&\color{blue}{189.64G}&841.57K&84.36\\
		FedBN-FT&208.99G&\underline{687.58K}&88.51&174.22G&610.31K&87.31&118.16G&407.71K&86.87\\
		FedBN-FedOPT&451.07G&1.78M&88.25&179.28G&758.36K&86.22&109.73G&487.2K&84.04\\
		FedBN-FedOPT-FT&\underline{195.3G}&696.81K&88.14&217.59G&762.72K&87.91&171.93G&593.18K&87.34\\
		\midrule
		Ditto&2.04T&1.78M&\color{red}{88.39}&700.28G&700.96K&\color{red}{88.87}&399.96G&533.77K&\color{red}{\underline{88.90}}\\
		Ditto-FT&2.73T&2.26M&85.72&349.06G&600.22K&67.97&433.51G&749.29K&72.08\\
		Ditto-FedBN&1.51T&1.02M&\textbf{88.94}&720.27G&623.37K&\textbf{89.33}&336.44G&695.85K&67.63\\
		Ditto-FedBN-FT&2.86T&1.89M&86.53&1.42T&1.19M&86.35&410.03G&762.09K&72.84\\
		Ditto-FedBN-FedOpt&2.45T&1.65M&\underline{88.73}&922.71G&801.91K&\underline{89.30}&408.13G&493.82K&\textbf{88.95}\\
		Ditto-FedBN-FedOpt-FT&2.52T&1.66M&87.02&1.42T&1.19M&87.34&876.33G&993.59K&79.38\\
		\midrule
		FedEM&2.98T&2.0M&84.35&1.99T&1.57M&84.49&1017.32G&1.03M&84.46\\
		FedEM-FT&5.79T&1.67M&86.17&4.12T&1.16M&86.13&3.83T&1.35M&86.11\\
		FedEM-FedBN&10.34T&3.1M&84.37&1.87T&1.26M&84.83&1.23T&1.15M&84.26\\
		FedEM-FedBN-FT&7.55T&2.59M&88.29&4.07T&1.34M&87.49&5.03T&1.29M&86.54\\
		FedEM-FedBN-FedOPT&6.0T&1.79M&82.12&3.89T&2.63M&85.81&1.81T&1.7M&85.16\\
		FedEM-FedBN-FedOPT-FT&13.86T&4.76M&87.54&4.11T&1.36M&87.69&3.62T&1.15M&85.91\\
		\bottomrule
	\end{tabular}
	\label{tab:compute_femnist}
\end{sidewaystable}

\begin{table}
	\centering
	\caption{The convergence results including the convergence round $T'$ and $\overline{Acc}$ for FEMNIST datasets with different $s$. The $T'=0$  indicates the early-stopping is not triggered within the large number of FL rounds $T=1000$.}
	\begin{tabular}{lllllll}
		\toprule
		& \multicolumn{2}{c}{FEMNIST, $s=0.2$} & \multicolumn{2}{c}{FEMNIST, $s=0.1$} & \multicolumn{2}{c}{FEMNIST, $s=0.05$} \\
		& $T'$&  $\overline{Acc}$&  
		$T'$ &  $\overline{Acc}$&
		$T'$ &  $\overline{Acc}$ \\
		\midrule
		FedAvg&\color{red}{173.33}&82.40&\color{red}{\underline{246.67}}&82.07&\color{blue}{350.00}&82.31\\
		FedAvg-FT&220.00&85.17&416.67&84.26&476.67&83.45\\
		FedOpt&733.33&19.59&420.00&27.47&\underline{73.33}&22.45\\
		FedOpt-FT&\underline{63.33}&10.39&\textbf{146.67}&30.88&800.00&30.02\\
		\midrule
		pFedMe&640.00&\color{blue}{86.50}&670.00&83.69&450.00&\color{blue}{84.29}\\
		pFedMe-FT&960.00&87.06&630.00&85.05&520.00&86.36\\
		\midrule
		FedBN&\color{blue}{273.33}&85.38&390.00&\color{blue}{83.81}&846.67&82.76\\
		FedBN-FT&340.00&\underline{87.65}&466.67&86.22&410.00&86.01\\
		FedBN-FedOPT&943.33&87.27&580.00&85.09&490.00&82.63\\
		FedBN-FedOPT-FT&360.00&87.13&583.33&87.10&596.67&86.53\\
		\midrule
		Ditto&406.67&\color{red}{87.18}&463.33&\color{red}{87.65}&486.67&\color{red}{\underline{87.80}}\\
		Ditto-FT&286.67&84.30&396.67&65.86&483.33&70.20\\
		Ditto-FedBN&536.67&\textbf{87.82}&476.67&\textbf{88.28}&700.00&66.15\\
		Ditto-FedBN-FT&\textbf{0.00}&85.16&263.33&84.98&766.67&71.76\\
		Ditto-FedBN-FedOpt&873.33&87.64&613.33&\underline{88.20}&496.67&\textbf{87.84}\\
		Ditto-FedBN-FedOpt-FT&880.00&85.71&263.33&86.00&\textbf{0.00}&76.75\\
		\midrule
		FedEM&290.00&82.61&\color{blue}{373.33}&83.13&\color{red}{346.67}&82.91\\
		FedEM-FT&243.33&84.91&276.67&84.77&453.33&84.61\\
		FedEM-FedBN&570.00&82.94&350.00&83.35&430.00&82.98\\
		FedEM-FedBN-FT&476.67&87.09&373.33&86.56&483.33&85.37\\
		FedEM-FedBN-FedOPT&330.00&80.48&730.00&85.01&633.33&83.87\\
		FedEM-FedBN-FedOPT-FT&876.67&86.23&376.67&86.58&430.00&84.98\\
		\bottomrule
	\end{tabular}
	\label{tab:converge_femnist}
\end{table}

\begin{sidewaystable}
	\centering
	\caption{The efficiency-accuracy trade-off results including total FLOPS, communication bytes (Com.), and $\overline{Acc}$ for CIFAR10 datasets with different $\alpha$.}
	\begin{tabular}{l|lll|lll|lll}
		\toprule
		& \multicolumn{3}{c|}{CIFAR10, $\alpha=5$} & \multicolumn{3}{c|}{CIFAR10, $\alpha=0.5$} & \multicolumn{3}{c}{CIFAR10, $\alpha=0.1$} \\
		& FLOPS & Com. &  $\overline{Acc}$&  
		FLOPS & Com. &  $\overline{Acc}$&
		FLOPS & Com. &  $\overline{Acc}$ \\
		\midrule
		FedAvg&\color{red}{\textbf{388.27G}}&654.43K&\color{blue}{73.86}&\color{blue}{384.14G}&\color{blue}{646.69K}&\color{red}{\underline{70.82}}&\color{red}{305.15G}&513.76K&\color{blue}{56.21}\\
		FedAvg-FT&422.66G&685.52K&73.89&404.17G&654.46K&70.12&\underline{259.72G}&425.4K&53.99\\
		\midrule
		FedOpt&1.26T&862.74K&54.66&494.38G&833.2K&49.90&651.29G&1.06M&36.41\\
		FedOpt-FT&1.37T&917.13K&57.33&1003.55G&1.59M&51.70&1.36T&895.36K&41.56\\
		\midrule
		pFedMe&3.26T&2.07M&\color{red}{\underline{73.92}}&3.95T&918.26K&70.30&2.51T&594.98K&55.70\\
		pFedMe-FT&3.57T&2.22M&\textbf{77.73}&3.57T&825.05K&\textbf{72.76}&4.15T&969.52K&\textbf{58.65}\\
		\midrule
		FedBN&\color{red}{\textbf{388.27G}}&\color{blue}{540.02K}&72.24&\color{red}{375.0G}&\color{red}{520.8K}&68.32&\color{blue}{632.15G}&\color{red}{350.44K}&\color{red}{56.36}\\
		FedBN-FT&\underline{394.14G}&527.19K&72.55&\underline{356.18G}&475.9K&67.45&\textbf{249.57G}&\underline{337.61K}&50.52\\
		FedBN-FedOPT&581.98G&809.39K&71.63&581.92G&809.43K&67.13&660.49G&360.45K&\underline{57.17}\\
		FedBN-FedOPT-FT&638.27G&854.28K&72.32&\textbf{288.66G}&\textbf{386.11K}&67.72&381.03G&\textbf{206.52K}&49.35\\
		\midrule
		Ditto&\color{blue}{1.81T}&\color{red}{\textbf{491.25K}}&72.02&1.31T&685.55K&67.96&929.51G&\color{blue}{479.8K}&45.67\\
		Ditto-FT&4.24T&1.12M&68.59&1.22T&631.15K&58.52&1001.19G&510.88K&37.79\\
		Ditto-FedBN&2.21T&\underline{495.12K}&71.70&1.87T&\underline{418.18K}&62.22&940.86G&392.52K&41.59\\
		Ditto-FedBN-FT&5.71T&1.24M&67.90&1.3T&552.87K&57.14&953.17G&392.52K&37.21\\
		Ditto-FedBN-FedOpt&2.33T&520.77K&71.58&1.68T&552.87K&60.78&1.36T&578.53K&47.94\\
		Ditto-FedBN-FedOpt-FT&3.09T&687.53K&68.41&1.01T&431.01K&56.66&1.18T&495.14K&38.73\\
		\midrule
		FedEM&6.78T&1.41M&73.34&7.42T&1.55M&\color{blue}{70.56}&5.74T&1.21M&53.43\\
		FedEM-FT&12.49T&1.81M&72.89&12.05T&1.74M&65.47&9.75T&1.45M&44.98\\
		FedEM-FedBN&6.88T&1.18M&71.43&8.58T&1.46M&68.42&7.61T&741.57K&55.30\\
		FedEM-FedBN-FT&9.91T&1.17M&70.46&11.44T&1.35M&61.43&9.01T&704.68K&43.48\\
		FedEM-FedBN-FedOPT&11.2T&1.91M&71.43&12.79T&2.18M&67.11&10.63T&1.01M&56.87\\
		FedEM-FedBN-FedOPT-FT&9.45T&1.12M&71.32&8.85T&1.05M&62.03&9.6T&758.5K&41.10\\
		\bottomrule
	\end{tabular}
	\label{tab:compute_cifar}
\end{sidewaystable}

\begin{table}
	\centering
	\caption{The convergence results including the convergence round $T'$ and $\overline{Acc}$ for CIFAR10 datasets with different $\alpha$. The $T'=0$  indicates the early-stopping is not triggered within the large number of FL rounds $T=1000$.}
	\begin{tabular}{l|ll|ll|ll}
		\toprule
		& \multicolumn{2}{c|}{CIFAR10, $\alpha=5$} & \multicolumn{2}{c|}{CIFAR10, $\alpha=0.5$} & \multicolumn{2}{c}{CIFAR10, $\alpha=0.1$} \\
		& $T'$&  $\overline{Acc}$&  
		$T'$ &  $\overline{Acc}$&
		$T'$ &  $\overline{Acc}$ \\
		\midrule
		FedAvg&280.00&\color{red}{\underline{73.87}}&276.67&\color{red}{\underline{70.89}}&213.33&\color{red}{57.71}\\
		FedAvg-FT&293.33&73.86&280.00&69.94&173.33&54.47\\
		FedOpt&370.00&54.66&356.67&49.61&466.67&37.30\\
		FedOpt-FT&393.33&57.35&696.67&51.69&383.33&42.01\\
		\midrule
		pFedMe&573.33&\color{blue}{73.74}&393.33&\color{blue}{70.63}&243.33&56.46\\
		pFedMe-FT&643.33&\textbf{77.82}&353.33&\textbf{73.46}&400.00&\textbf{59.74}\\
		\midrule
		FedBN&280.00&72.17&\color{blue}{270.00}&68.28&\color{red}{173.33}&\color{blue}{57.59}\\
		FedBN-FT&273.33&72.50&246.67&67.69&166.67&51.40\\
		FedBN-FedOPT&420.00&71.59&420.00&67.22&186.67&58.19\\
		FedBN-FedOPT-FT&443.33&72.27&\underline{200.00}&68.07&\textbf{106.67}&50.76\\
		\midrule
		Ditto&\color{red}{\underline{210.00}}&71.97&293.33&67.65&\color{blue}{196.67}&49.77\\
		Ditto-FT&490.00&68.57&270.00&58.63&210.00&38.97\\
		Ditto-FedBN&256.67&71.65&216.67&62.65&203.33&43.07\\
		Ditto-FedBN-FT&660.00&67.78&286.67&57.27&203.33&39.13\\
		Ditto-FedBN-FedOpt&270.00&71.54&286.67&61.12&300.00&48.27\\
		Ditto-FedBN-FedOpt-FT&356.67&68.37&223.33&57.59&256.67&40.67\\
		\midrule
		FedEM&\color{blue}{213.33}&73.36&\color{red}{233.33}&70.56&\color{red}{173.33}&54.52\\
		FedEM-FT&273.33&72.84&263.33&65.49&210.00&42.92\\
		FedEM-FedBN&216.67&71.40&270.00&68.64&133.33&57.82\\
		FedEM-FedBN-FT&216.67&70.40&250.00&62.13&\underline{126.67}&43.97\\
		FedEM-FedBN-FedOPT&353.33&71.39&403.33&67.19&186.67&\underline{58.40}\\
		FedEM-FedBN-FedOPT-FT&\textbf{206.67}&71.27&\textbf{193.33}&62.31&136.67&40.70\\
		\bottomrule
	\end{tabular}
	\label{tab:converge_cifar}
\end{table}

\begin{sidewaystable}
	\centering
	\caption{The efficiency-accuracy trade-off results including total FLOPS, communication bytes (Com.), and $\overline{Acc}$ for COLA, SST-2 and Twitter datasets.}
	\begin{tabular}{l|lll|lll|lll}
		\toprule
		& \multicolumn{3}{c|}{COLA} & \multicolumn{3}{c}{SST-2} & \multicolumn{3}{c}{Twitter} \\
		& FLOPS & Com. &  $\overline{Acc}$&  
		FLOPS & Com. &  $\overline{Acc}$&  
		FLOPS & Com. &  $\overline{Acc}$\\
		\midrule
		FedAvg&\color{red}{\textbf{22.15G}}&\color{blue}{310.94K}&\color{blue}{71.85}&\color{red}{954.66G}&\color{blue}{464.37K}&\color{blue}{74.88}&\color{blue}{\underline{726.45K}}&60.32K&62.15\\
		FedAvg-FT&\underline{37.23G}&\underline{251.93K}&68.29&1.41T&629.61K&74.14&4.75M&19.26K&70.53\\
		FedOpt&50.71G&299.14K&71.85&808.78G&393.56K&72.28&\underline{726.45K}&60.32K&62.09\\
		FedOpt-FT&58.3G&393.56K&62.59&\underline{707.21G}&310.94K&65.77&4.52M&\underline{18.35K}&\underline{71.08}\\
		\midrule
		pFedMe&208.22G&1.04M&\color{red}{\underline{74.40}}&4.66T&830.51K&71.27&\color{red}{\textbf{603.46K}}&\color{red}{29.3K}&\color{blue}{63.45}\\
		pFedMe-FT&240.44G&357.98K&\textbf{78.47}&3.76T&629.4K&\underline{75.61}&12.07M&\textbf{15.57K}&\textbf{84.00}\\
		\midrule
		FedBN&\color{blue}{46.07G}&\color{red}{259.58K}&\color{blue}{71.85}&\color{red}{954.66G}&\color{blue}{464.37K}&\color{blue}{74.88}&-&-&-\\
		FedBN-FT&40.29G&259.58K&66.71&1.06T&476.17K&68.81&-&-&-\\
		FedBN-FedOPT&86.98G&482.11K&71.85&\textbf{593.16G}&\underline{270.71K}&64.70&-&-&-\\
		FedBN-FedOPT-FT&39.08G&\textbf{248.45K}&67.48&\underline{707.21G}&292.96K&68.65&-&-&-\\
		\midrule
		Ditto&133.52G&322.75K&55.46&\color{blue}{1.43T}&\color{red}{275.53K}&52.03&2.64M&\color{blue}{38.42K}&\color{red}{70.23}\\
		Ditto-FT&124.81G&440.77K&72.11&1.85T&334.55K&56.49&4.2M&36.59K&69.99\\
		Ditto-FedBN&96.29G&404.22K&70.69&1.48T&\underline{270.71K}&56.03&-&-&-\\
		Ditto-FedBN-FT&113.99G&381.97K&72.66&1.58T&\textbf{270.7K}&53.15&-&-&-\\
		Ditto-FedBN-FedOpt&163.88G&370.84K&50.25&3.19T&626.86K&57.67&-&-&-\\
		Ditto-FedBN-FedOpt-FT&148.0G&292.96K&55.01&1.58T&537.74K&52.89&-&-&-\\
		\midrule
		FedEM&414.28G&801.7K&\color{blue}{71.85}&18.46T&1.55M&\color{red}{\textbf{75.78}}&18.82M&95.64K&63.44\\
		FedEM-FT&1.85T&1.02M&54.90&24.19T&1.32M&64.86&24.19M&35.27K&70.97\\
		FedEM-FedBN&729.78G&753.83K&71.44&16.84T&1.31M&75.43&-&-&-\\
		FedEM-FedBN-FT&1.37T&979.64K&57.62&24.19T&1.24M&64.96&-&-&-\\
		FedEM-FedBN-FedOPT&414.28G&753.83K&71.85&13.34T&1.05M&72.25&-&-&-\\
		FedEM-FedBN-FedOPT-FT&1.49T&1.02M&57.23&22.0T&1.12M&62.26&-&-&-\\
		\bottomrule
	\end{tabular}
	\label{tab:compute_nlp}
\end{sidewaystable}

\begin{table}
	\centering
	\caption{The convergence results including the convergence round $T'$ and $\overline{Acc}$ for COLA, SST-2 and Twitter datasets. The $T'=0$ indicates the early-stopping is not triggered within a large number of FL rounds, $T=500$ for COLA and SST-2, and $T=1000$ for Twitter datasets.}
	\begin{tabular}{l|ll|ll|ll}
		\toprule
		& \multicolumn{2}{c|}{COLA} & \multicolumn{2}{c}{SST-2} & \multicolumn{2}{c}{Twitter}\\
		& $T'$&  $\overline{Acc}$&  
		$T'$ &  $\overline{Acc}$ &  
		$T'$ &  $\overline{Acc}$\\
		\midrule
		FedAvg&\color{blue}{43.33}&51.53&\color{blue}{65.00}&\color{blue}{\underline{76.30}}&223.33&56.98\\
		FedAvg-FT&\textbf{35.00}&58.74&88.33&75.36&70.67&68.45\\
		FedOpt&41.67&57.10&55.00&73.78&220.33&59.95\\
		FedOpt-FT&55.00&59.77&43.33&66.17&66.67&69.20\\
		\midrule
		pFedMe&150.00&\color{red}{\underline{67.58}}&116.67&65.08&\color{red}{106.67}&58.18\\
		pFedMe-FT&50.00&\textbf{69.17}&88.33&74.36&\underline{53.33}&\textbf{78.82}\\
		\midrule
		FedBN&\color{red}{38.33}&\color{blue}{59.60}&\color{blue}{65.00}&\color{blue}{\underline{76.30}}&-&-\\
		FedBN-FT&38.33&59.69&66.67&68.50&-&-\\
		FedBN-FedOPT&71.67&59.60&\underline{40.00}&65.59&-&-\\
		FedBN-FedOPT-FT&\underline{36.67}&59.10&43.33&68.42&-&-\\
		\midrule
		Ditto&45.00&55.14&\color{red}{\textbf{38.33}}&49.94&\color{blue}{140.33}&\color{red}{66.90}\\
		Ditto-FT&61.67&63.61&46.67&54.34&133.33&66.91\\
		Ditto-FedBN&60.00&62.68&\underline{40.00}&49.44&-&-\\
		Ditto-FedBN-FT&56.67&63.58&\underline{40.00}&52.18&-&-\\
		Ditto-FedBN-FedOpt&55.00&52.48&93.33&55.61&-&-\\
		Ditto-FedBN-FedOpt-FT&43.33&57.13&80.00&53.16&-&-\\
		\midrule
		FedEM&\color{red}{38.33}&51.52&76.67&\color{red}{\textbf{76.53}}&163.33&\color{blue}{61.70}\\
		FedEM-FT&50.00&57.80&65.00&64.29&\textbf{40.67}&\underline{70.19}\\
		FedEM-FedBN&38.33&57.95&68.33&75.06&-&-\\
		FedEM-FedBN-FT&50.00&58.74&65.00&64.33&-&-\\
		FedEM-FedBN-FedOPT&38.33&59.60&55.00&72.66&-&-\\
		FedEM-FedBN-FedOPT-FT&53.33&56.74&58.33&58.42&-&-\\
		\bottomrule
	\end{tabular}
	\label{tab:converge_nlp}
\end{table}

\begin{table}
	\centering
	\caption{The convergence results including the convergence round $T'$ and $\overline{Loss}$ for Movielens-1M and Movielens-10M datasets. The $T'=0$ indicates the early-stopping is not triggered within the large number of FL rounds $T=1000$.}
	\begin{tabular}{l|ll|ll}
		\toprule
		& \multicolumn{2}{c|}{Movielens-1M} & \multicolumn{2}{c}{Movielens-10M} \\
		& $T'$&  $\overline{Loss}$&  
		$T'$ &  $\overline{Loss}$ \\
		\midrule
		FedAvg&\color{red}{360.33}&\color{blue}{0.85}&\color{blue}{\underline{470.67}}&\color{red}{\textbf{0.71}}\\
		FedAvg-FT&0&0.85&520.33&\textbf{0.71}\\
		FedAvg-FT-FedOpt&830.0&0.86&0&0.75\\
		FedOpt&\textbf{270.0}&0.84&0&\underline{0.72}\\
		FedOpt-FT&\underline{300.0}&0.84&0&0.77\\
		\midrule
		pFedMe&0&\color{red}{\textbf{0.55}}&\color{red}{\textbf{280.33}}&12.48\\
		pFedMe-FT&470.0&\underline{0.60}&840.00&0.80\\
		\midrule
		Ditto&\color{red}{360.67}&1.31&910.67&\color{blue}{1.81}\\
		Ditto-FT&450.33&1.35&0&2.30\\
		Ditto-FT-FedOpt&550.67&1.35&0&1.98\\
		\midrule
		FedEM&\color{blue}{700.33}&0.87&0&2.37\\
		FedEM-FT&780.00&0.87&0&0.98\\
		FedEM-FT-FedOpt&0&0.87&0&1.88\\
		\bottomrule
	\end{tabular}
	\label{tab:converge_rec}
\end{table}

\begin{sidewaystable}
	\centering
	\caption{The efficiency-accuracy trade-off results including total FLOPS, communication bytes (Com.), and $\overline{Acc}$ for Pubmed, Cora, and Citeseer datasets.}
	\begin{tabular}{l|lll|lll|lll}
		\toprule
		& \multicolumn{3}{c|}{PUBMED} & \multicolumn{3}{c|}{CORA} & \multicolumn{3}{c}{CITESEER} \\
		& FLOPS & Com. &  $\overline{Acc}$&  
		FLOPS & Com. &  $\overline{Acc}$&
		FLOPS & Com. &  $\overline{Acc}$ \\
		\midrule
		FedAvg&\color{blue}{40.41G}&909.81K&\color{blue}{87.27}&\color{blue}{2.94G}&980.35K&81.30&\color{blue}{71.06G}&2.98M&\color{blue}{75.58}\\
		FedAvg-FT&31.39G&1.79M&87.21&5.16G&885.92K&82.07&77.54G&2.39M&75.63\\
		FedOpt&163.3G&6.92M&67.38&\underline{2.22G}&744.07K&70.70&\textbf{3.89G}&\underline{437.18K}&71.59\\
		FedOpt-FT&24.78G&791.28K&82.36&6.27G&1.05M&82.68&25.07G&791.28K&74.34\\
		\midrule
		pFedMe&45.67G&1.0M&86.91&10.02G&1.26M&83.18&193.83G&\color{blue}{744.07K}&75.30\\
		pFedMe-FT&119.12G&862.1K&85.71&22.76G&2.09M&82.11&64.31G&460.79K&75.35\\
		\midrule
		FedBN&\color{red}{27.53G}&\color{red}{875.47K}&\color{red}{\textbf{88.49}}&\color{red}{2.51G}&\color{red}{\underline{615.38K}}&\color{red}{84.13}&\color{red}{\underline{14.01G}}&\color{red}{441.99K}&\color{red}{75.80}\\
		FedBN-FT&35.86G&2.04M&87.45&29.13G&4.85M&76.20&28.25G&531.61K&75.07\\
		FedBN-FedOPT&\textbf{13.26G}&\textbf{632.72K}&87.87&\textbf{2.07G}&736.76K&84.64&26.08G&823.45K&76.20\\
		FedBN-FedOPT-FT&45.55G&1.04M&87.54&3.21G&\underline{615.38K}&84.10&21.97G&771.43K&\textbf{76.70}\\
		\midrule
		Ditto&46.9G&\color{blue}{909.53K}&\color{blue}{87.27}&7.27G&\color{blue}{933.13K}&\color{blue}{83.67}&79.77G&838.49K&74.79\\
		Ditto-FT&43.73G&1.54M&87.47&8.59G&909.31K&81.47&42.34G&744.07K&76.47\\
		Ditto-FedBN&\underline{23.72G}&752.63K&\underline{88.18}&8.37G&788.77K&81.38&37.51G&528.68K&75.35\\
		Ditto-FedBN-FT&36.49G&963.57K&87.83&5.46G&\textbf{424.65K}&83.25&26.06G&494.0K&75.97\\
		Ditto-FedBN-FedOpt&57.84G&821.99K&87.81&7.45G&702.08K&82.54&20.14G&\textbf{407.31K}&75.07\\
		Ditto-FedBN-FedOpt-FT&49.88G&\underline{650.06K}&87.60&7.92G&\underline{615.38K}&82.00&35.48G&684.74K&76.42\\
		\midrule
		FedEM&575.33G&3.24M&85.64&61.17G&2.01M&81.92&760.69G&8.36M&75.41\\
		FedEM-FT&647.22G&3.04M&85.88&61.14G&2.7M&77.32&150.55G&2.22M&72.71\\
		FedEM-FedBN&253.94G&2.07M&88.12&18.99G&1.92M&\textbf{85.07}&52.64G&1.82M&74.90\\
		FedEM-FedBN-FT&275.77G&3.02M&86.38&34.58G&1.72M&84.61&89.54G&1.47M&75.29\\
		FedEM-FedBN-FedOPT&229.51G&1.87M&87.56&108.73G&2.62M&84.68&419.47G&3.37M&76.08\\
		FedEM-FedBN-FedOPT-FT&233.63G&2.12M&87.49&285.87G&4.92M&\underline{85.02}&561.46G&3.22M&\underline{76.59}\\
		\bottomrule
	\end{tabular}
	\label{tab:compute_graph}
\end{sidewaystable}

\begin{table}
	\centering
	\caption{The convergence results including the convergence round $T'$ and $\overline{Acc}$ for Pubmed, Cora, and Citeseer datasets. The $T'$ indicates the early-stopping is not triggered within the large number of FL rounds $T=500$.}
	\begin{tabular}{l|ll|ll|ll}
		\toprule
		& \multicolumn{2}{c|}{PUBMED} & \multicolumn{2}{c|}{CORA} & \multicolumn{2}{c}{CITESEER} \\
		& $T'$&  $\overline{Acc}$&  
		$T'$ &  $\overline{Acc}$&
		$T'$ &  $\overline{Acc}$ \\
		\midrule
		FedAvg&\color{red}{63.33}&86.72&68.33&81.07&215.00&\color{red}{75.64}\\
		FedAvg-FT&128.33&86.71&61.67&81.90&171.67&75.77\\
		FedOpt&\textbf{0.00}&66.69&51.67&70.17&\textbf{30.00}&71.60\\
		FedOpt-FT&\underline{55.00}&81.53&75.00&82.31&55.00&74.38\\
		\midrule
		pFedMe&\color{blue}{71.67}&86.35&90.00&82.76&\color{blue}{51.67}&75.36\\
		pFedMe-FT&60.00&85.47&150.00&81.98&\underline{31.67}&75.40\\
		\midrule
		FedBN&83.33&\color{red}{\textbf{87.97}}&\color{blue}{58.33}&\color{red}{83.64}&\color{red}{41.67}&\color{blue}{75.59}\\
		FedBN-FT&146.67&87.02&183.33&76.01&36.67&75.16\\
		FedBN-FedOPT&60.00&87.43&70.00&84.11&78.33&76.33\\
		FedBN-FedOPT-FT&101.67&87.02&58.33&83.79&73.33&\textbf{76.77}\\
		\midrule
		Ditto&\color{red}{63.33}&\color{blue}{86.85}&65.00&\color{blue}{83.50}&58.33&75.55\\
		Ditto-FT&110.00&87.10&63.33&81.53&51.67&76.57\\
		Ditto-FedBN&71.67&\underline{87.75}&75.00&81.36&50.00&75.46\\
		Ditto-FedBN-FT&91.67&87.43&\textbf{40.00}&82.21&46.67&76.05\\
		Ditto-FedBN-FedOpt&78.33&87.27&66.67&82.38&38.33&75.19\\
		Ditto-FedBN-FedOpt-FT&61.67&87.10&58.33&81.99&65.00&76.52\\
		\midrule
		FedEM&78.33&85.05&\color{red}{\underline{48.33}}&81.72&203.33&75.49\\
		FedEM-FT&73.33&85.54&65.00&78.43&53.33&72.88\\
		FedEM-FedBN&68.33&87.63&63.33&\underline{84.45}&60.00&74.29\\
		FedEM-FedBN-FT&100.00&85.68&56.67&\textbf{84.57}&48.33&75.40\\
		FedEM-FedBN-FedOPT&61.67&87.11&86.67&83.88&111.67&76.17\\
		FedEM-FedBN-FedOPT-FT&70.00&87.16&163.33&84.40&106.67&\underline{76.74}\\
		\bottomrule
	\end{tabular}
	\label{tab:converge_graph}
\end{table}

\begin{table}
	\centering
	\caption{The efficiency-accuracy trade-off results including total FLOPS, communication bytes (Com.), and $\overline{Loss}$ for Movielens-1M and Movielens-10M datasets.}
	\begin{tabular}{l|lll|lll}
		\toprule
		& \multicolumn{3}{c|}{Movielens-1M} & \multicolumn{3}{c}{Movielens-10M}  \\
		& FLOPS & Com. &  $\overline{Loss}$&  
		FLOPS & Com. &  $\overline{Loss}$\\
		\midrule
		FedAvg&\color{blue}{343.73M}&\color{red}{108.75K}&\color{blue}{0.84}&375.55G&142.58K&\color{red}{\textbf{0.70}}\\
		FedAvg-FT&995.55M&301.5K&0.84&162.79G&157.72K&\underline{0.71}\\
		FedAvg-FT-FedOpt&\underline{236.68M}&250.45K&0.85&\underline{21.39G}&302.91K&0.73\\
		FedOpt&258.31M&\textbf{81.62K}&0.83&\textbf{19.75G}&302.91K&\underline{0.71}\\
		FedOpt-FT&299.3M&\underline{90.66K}&0.83&52.04G&302.91K&0.74\\
		\midrule
		pFedMe&763.2M&\color{blue}{301.51K}&\color{red}{\textbf{0.54}}&131.49G&\color{red}{\textbf{24.44K}}&13.06\\
		pFedMe-FT&376.37M&142.35K&\underline{0.60}&93.58G&254.7K&0.80\\
		\midrule
		Ditto&\color{red}{332.94M}&\color{red}{108.75K}&1.29&\color{red}{72.4G}&302.91K&1.84\\
		Ditto-FT&\textbf{220.62M}&135.88K&1.35&242.74G&302.91K&1.69\\
		Ditto-FT-FedOpt&269.63M&166.03K&1.36&73.26G&302.91K&2.03\\
		\midrule
		FedEM&2.07G&523.1K&0.85&\color{blue}{86.69G}&\color{blue}{135.15K}&\color{blue}{1.75}\\
		FedEM-FT&4.07G&582.82K&0.85&253.37G&269.78K&0.87\\
		FedEM-FT-FedOpt&5.22G&746.53K&0.86&113.27G&\underline{120.19K}&1.43\\
		\bottomrule
	\end{tabular}
	\label{tab:compute_rec}
\end{table}

\begin{table}[!t]
	\centering
	
	\caption{Efficiency results in terms of process memory (MB) and running time (seconds). A higher value indicates that more system resources were consumed. The $MEM_{avg}$ and $MEM_{peak}$ indicates average and peak values of process-used memory respectively, and $T_{run}$ indicates the process running time. Similar to above table, \textbf{Bold} and \underline{underlined} indicate the best and second-best results among all compared methods, while \textcolor{red}{red} and \textcolor{blue}{blue} indicate the best and second-best results for original methods without combination ``-''.
	}
	\vspace{-0.05in}
	\small
	\resizebox{\columnwidth}{!}{
		\begin{tabular}{p{1.42in}|ccc|ccc}
			\toprule
			& \multicolumn{3}{c|}{FEMNIST, $s=0.2$} & \multicolumn{3}{c}{SST-2} \\
			& $MEM_{avg}$ & $MEM_{peak}$ & $T_{run}$ & $MEM_{avg}$ & $MEM_{peak}$ & $T_{run}$ \\
			\midrule
Global-Train&\color{red}{\textbf{86.56}}&\color{red}{\textbf{86.56}}&\color{red}{\textbf{11.00}}&\color{red}{\textbf{87.23}}&\color{red}{\textbf{93.37}}&892.00\\
Isolated&\color{blue}{746.85}&1351.94&1108.00&\color{blue}{\underline{118.00}}&\color{blue}{\underline{151.07}}&994.00\\
FedAvg&10506.71&17707.04&\color{blue}{\underline{840.00}}&297.43&399.51&\color{red}{319.00}\\
FedAvg-FT&13400.46&21752.04&1193.00&314.05&371.73&545.00\\
			\midrule
FedProx&19389.09&20729.26&1415.00&6635.17&7378.15&672.00\\
FedProx-FT&21209.42&22503.71&1935.00&7804.79&8160.99&1810.00\\
			\midrule
pFedMe&1880.78&1979.59&7205.00&262.67&366.15&1572.00\\
pFedMe-FT&18573.50&21332.41&4676.00&282.19&367.81&1080.00\\
HypCluster&21659.61&22386.94&2803.00&6942.08&7510.02&678.00\\
HypCluster-FT&22569.13&23588.59&3602.00&7624.09&8232.55&887.00\\
			\midrule
FedBN&11135.25&15406.92&1011.00&279.89&373.27&\color{blue}{332.00}\\
FedBN-FT&17347.31&25649.89&936.00&266.80&331.84&551.00\\
FedBN-FedOPT&21866.64&35335.47&2881.00&216.22&311.33&\textbf{202.00}\\
FedBN-FedOPT-FT&9827.22&14711.30&845.00&260.59&350.32&\underline{283.00}\\
			\midrule
Ditto&1127.92&1268.18&4628.00&149.16&204.95&638.00\\
Ditto-FT&1110.87&1532.88&8915.00&228.51&292.20&599.00\\
Ditto-FedBN&1116.05&1227.17&4049.00&196.52&228.59&593.00\\
Ditto-FedBN-FT&1453.92&1730.06&8528.00&216.48&290.13&627.00\\
Ditto-FedBN-FedOpt&1176.76&1273.67&5782.00&202.71&237.94&560.00\\
Ditto-FedBN-FedOpt-FT&1299.42&1564.89&7545.00&275.08&333.58&637.00\\
			\midrule
FedEM&939.58&\color{blue}{1063.35}&5070.00&267.37&357.79&1568.00\\
FedEM-FT&563.77&619.02&7592.00&327.36&374.10&3210.00\\
FedEM-FedBN&572.14&912.14&12051.00&269.62&337.22&1615.00\\
FedEM-FedBN-FT&\underline{489.82}&\underline{616.71}&9391.00&306.51&367.51&3040.00\\
FedEM-FedBN-FedOPT&755.61&842.33&8977.00&251.02&343.16&1314.00\\
FedEM-FedBN-FedOPT-FT&608.29&638.73&19137.00&300.92&353.46&3116.00\\
			\bottomrule
	\end{tabular}}
	\label{tab:memory-time}
	\vspace{-0.05in}
\end{table}

\balance

\end{document}